\documentclass[journal]{IEEEtran}
\usepackage{lineno}
\modulolinenumbers[5]
\usepackage{xfrac}
\usepackage{changepage}
\usepackage{amsfonts}
\usepackage{amsmath}
\usepackage{graphicx}
\usepackage{subfig}
\usepackage{multirow}
\usepackage{array}
\newcolumntype{P}[1]{>{\centering\arraybackslash}p{#1}}
\usepackage{makecell}
\usepackage{pgfplots}
\usepackage[flushleft]{threeparttable}
\usepackage{alphalph}
\usepackage{bm}
\usepackage{hyperref}
\begin{document}
%
\title{Relation Network for Multi-label Aerial Image Classification}

\author{Yuansheng~Hua,~
        Lichao~Mou,~
        and~Xiao~Xiang~Zhu,~\IEEEmembership{Senior~Member,~IEEE}
\thanks{
This work is jointly supported by the China Scholarship Council, the European Research Council (ERC) under the European Unions Horizon 2020 research and innovation programme (grant agreement No [ERC-2016-StG-714087], Acronym: \textit{So2Sat}), and Helmholtz Association under the framework of the Young Investigators Group ``SiPEO'' (VH-NG-1018, \url{www.sipeo.bgu.tum.de}), Helmholtz Artificial Intelligence Cooperation Unit (HAICU) - Local Unit ``Munich Unit @Aeronautics, Space and Transport (MASTr)'' and Helmholtz Excellent Professorship ``Data Science in Earth Observation - Big Data Fusion for Urban Research''. Besides, the authors would like to thank Xinyi Liu for supporting this work with data annotation. \textit{(Corresponding author: Xiao Xiang Zhu.)}

Y. Hua, L. Mou, and X. X. Zhu are with the Remote Sensing Technology Institute, German Aerospace Center, 82234 Weßling, Germany, and also with Signal Processing in Earth Observation, Technical University of Munich, 80333 Munich, Germany (e-mail: yuansheng.hua@dlr.de; lichao.mou@dlr.de; xiaoxiang.zhu@dlr.de).}
}

\markboth{Submitted to IEEE TRANSACTIONS ON GEOSCIENCE AND REMOTE SENSING}%
{Shell \MakeLowercase{\textit{HUA et al.}}: Relation Network for Multi-label Aerial Image Classification}
%



\maketitle

\begin{abstract}

\textit{This is a preprint. To read the final version please visit IEEE Transactions on Geoscience and Remote Sensing.} Multi-label classification plays a momentous role in perceiving intricate contents of an aerial image and triggers several related studies over the last years. However, most of them deploy few efforts in exploiting label relations, while such dependencies are crucial for making accurate predictions. Although an LSTM layer can be introduced to modeling such label dependencies in a chain propagation manner, the efficiency might be questioned when certain labels are improperly inferred. To address this, we propose a novel aerial image multi-label classification network, attention-aware label relational reasoning network. Particularly, our network consists of three elemental modules: 1) a label-wise feature parcel learning module, 2) an attentional region extraction module, and 3) a label relational inference module. To be more specific, the label-wise feature parcel learning module is designed for extracting high-level label-specific features. The attentional region extraction module aims at localizing discriminative regions in these features without region proposal generation, and yielding attentional label-specific features. The label relational inference module finally predicts label existences using label relations reasoned from outputs of the previous module. The proposed network is characterized by its capacities of extracting discriminative label-wise features and reasoning about label relations naturally and interpretably. In our experiments, we evaluate the proposed model on two multi-label aerial image datasets, of which one is newly produced. Quantitative and qualitative results on these two datasets demonstrate the effectiveness of our model. To facilitate progress in the multi-label aerial image classification, our produced dataset will be made publicly available. 

\end{abstract}


\IEEEpeerreviewmaketitle

\section{Introduction} 

\begin{figure}[!t]
\centering
\subfloat[]{\includegraphics[width=.11\textwidth]{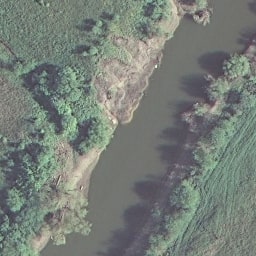}
\label{fig:ss1}}
\hfil
\subfloat[]{\includegraphics[width=.11\textwidth]{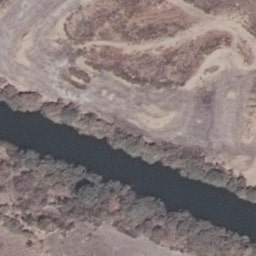}
\label{fig:ss2}}
\hfil
\subfloat[]{\includegraphics[width=.11\textwidth]{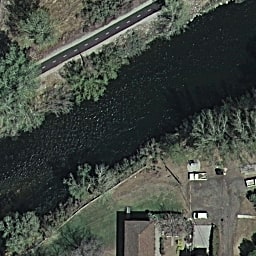}
\label{fig:ss3}}
\hfil
\subfloat[]{\includegraphics[width=.11\textwidth]{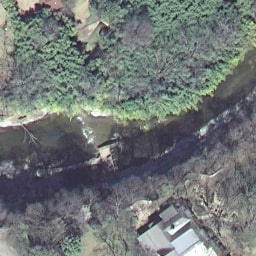}
\label{fig:ss4}}
\caption{Example aerial images of scene \texttt{river} and \textsl{objects} present in them. (a) \textsl{bare soil}, \textsl{grass}, \textsl{tree}, and \textsl{water}. (b) \textsl{water}, \textsl{bare soil}, and \textsl{tree}. (c) \textsl{water}, \textsl{building}, \textsl{grass}, \textsl{car}, \textsl{tree}, \textsl{pavement}, and \textsl{bare soil}. (d) \textsl{water}, \textsl{building}, \textsl{grass}, \textsl{bare soil}, \textsl{tree}, and \textsl{sand}.}
\label{fig:ss01}
\end{figure}

Recent advancements of remote sensing techniques have boosted the volume of attainable high-resolution aerial images, and massive amounts of applications, such as urban cartography \cite{Marmanis17, beyongrgb, Marcos18, mou2018rifcn}, traffic monitoring \cite{mou2018vehicle, 7729468, hsf}, terrain surface analysis \cite{lucchesi2013applications, Mou18, weng2018land, cheng2017remote}, and ecological scrutiny \cite{zarco2014tree, wen2017semantic}, have benefited from these developments. For this reason, the aerial image classification has become one of the fundamental visual tasks in the remote sensing community and drawn a plethora of research interests \cite{nogueira2017towards, zhu2017deep, 7358126, rs71114680, hu2018recent, zhang2015saliency, HUANG2018127, mou2017multitemporal}. The classification of aerial images refers to assigning these images with specific labels according to their semantic contents, and a common hypothesis shared by many relevant studies is that an image should be labeled with only one semantic category, such as scene categories (see Fig. \ref{fig:ss01}). Although such image-level labels \cite{yang2010bag, xia2017aid} are capable of delineating images from a macroscopic perspective, it is infeasible for them to provide a comprehensive view of objects in aerial images. To tackle this, huge quantities of algorithms have been proposed to identify each pixel in an image \cite{ren2015faster, long2015fully, badrinarayanan2015segnet} or localize objects with bounding boxes \cite{viola2001rapid, lin2017feature, ren2017object}. However, the acquisition of requisite ground truths (i.e., pixel-wise annotations and bounding boxes) demands enormous expertise and human labors, which makes relevant datasets expensive and difficult to access. With this intention, multi-label image classification now attracts increasing attention in the remote sensing community \cite{karalas2016land, zeggada2017deep, koda2018spatial, zeggada2018multilabel, hua2019recurrently} owing to that 1) a comprehensive picture of aerial image contents can be drawn, and 2) datasets required in this task are not expensive (only image-level labels are needed).

Fig.~\ref{fig:ss01} illustrates the difference between image-level scene labels and object labels. As shown in this figure, although these four images are assigned with the same scene label, their multiple object labels vary a lot. It is worth noting that the identification of some objects can actually offer important cues to understand a scene more deeply. For example, the existence of \textsl{building} and \textsl{pavement} indicates a high probability that rivers in Fig. \ref{fig:ss3} and \ref{fig:ss4} are very close to areas with frequent human activities, while rivers in Fig. \ref{fig:ss1} and \ref{fig:ss2} are more likely in the wild due to the absence of human activity cues. In contrast, simply recognizing scene labels can hardly provide such information. Therefore, in this paper, we dedicate our efforts to explore an effective model for the multi-label classification of aerial images.

\subsection{Challenges of Identifying Multiple labels}
\label{sec:challenge}

In identifying multiple labels of an aerial image, two main challenges need to be faced with. One is how to extract semantic feature representations from raw images. This is crucial but difficult especially for high-resolution aerial images, as they always contain complicated spatial contextual information.
Conventional approaches mainly resort to manually crafted features and semantic models \cite{yang2010bag, shao2013hierarchical, risojevic2013fusion, lowe2004distinctive, 7466064}, while these methods cannot effectively extract high-level semantics and lead to a limited performance in classification\cite{xia2017aid}. Hence an efficient high-level feature extractor is desirable.

The other challenge is how to take full advantage of label correlations to infer multiple object labels of an aerial image. In contrast to single-label classification, which mainly focuses on modeling image-label relevance, exploring and modeling label-label correlations plays a supplementary yet essential role in identifying multiple objects in aerial images. For instance, the presence of ships confidently infers the co-occurrence of water or sea, while the existence of a car suggests a high probability of the appearance of pavements. Unfortunately, such label correlations are scarcely addressed in the literature. One solution is to use a recurrent neural network (RNN) to learn label dependencies. However, this is done with a chain propagation fashion, and its performance heavily depends on the learning effectiveness of its long-term memorization. Moreover, in this way, label relations are modeled implicitly, which leads to a lack of interpretability.

Overall, an efficient multi-label classification model is supposed to be capable of not only learning high-level feature representations but also modeling label correlations effectively.

\subsection{Related Work}

Zegeye and Demir \cite{teshome2018novel} propose a multi-label active learning framework using a multi-label support vector machine (SVM), relying on both the multi-label uncertainty and diversity. Koda et al. \cite{koda2018spatial} introduce a spatial and structure SVM for multi-label classification by considering spatial relations between a given patch and its neighbors. Similarly, Zeggada et al. \cite{zeggada2018multilabel} employ a conditional random field (CRF) framework to model spatial contextual information among adjacent patches for improving the performance of classifying multiple object labels. 

With the development of computational resources and deep learning, very recent approaches mainly resort to deep networks for multi-label classification. In~\cite{zeggada2017deep}, the authors make use of a standard CNN architecture to extract feature representations and then feed them into a multi-label classification layer, which is composed of customized thresholding operations, for predicting multiple labels. In~\cite{stivaktakis2019deep}, the authors demonstrate that training a CNN for multi-label classification with a limited amount of labeled data usually leads to an underwhelming-performance model and propose a dynamic data augmentation method for enlarging training sets. More recently, Sumbul and Demir~\cite{Demir2019label} propose a CNN-RNN method for identifying labels in multi-spectral images, where a bidirectional LSTM is employed to model spatial relationships among image patches. In order to explore inherent correlations among object labels, \cite{hua2019recurrently} proposes a CNN-LSTM hybrid network architecture to learn label dependencies for classifying object labels of aerial images. Besides, we also notice that several zero short learning researches focus on employing prior knowledge to model label relations. For instance, Sumbul et al. \cite{sumbul2017fine} apply an unsupervised word embedding model to encoding labels into word vectors, which are supposed to contain label semantics, and then model label relationships with these vectors. Lee et al. \cite{lee2018multi} propose to learn label relations from structured knowledge graphs observed from the real world.

\subsection{The Motivation of Our Work}
\begin{figure*}[!t]
    \centering
    \includegraphics[width=0.95\textwidth]{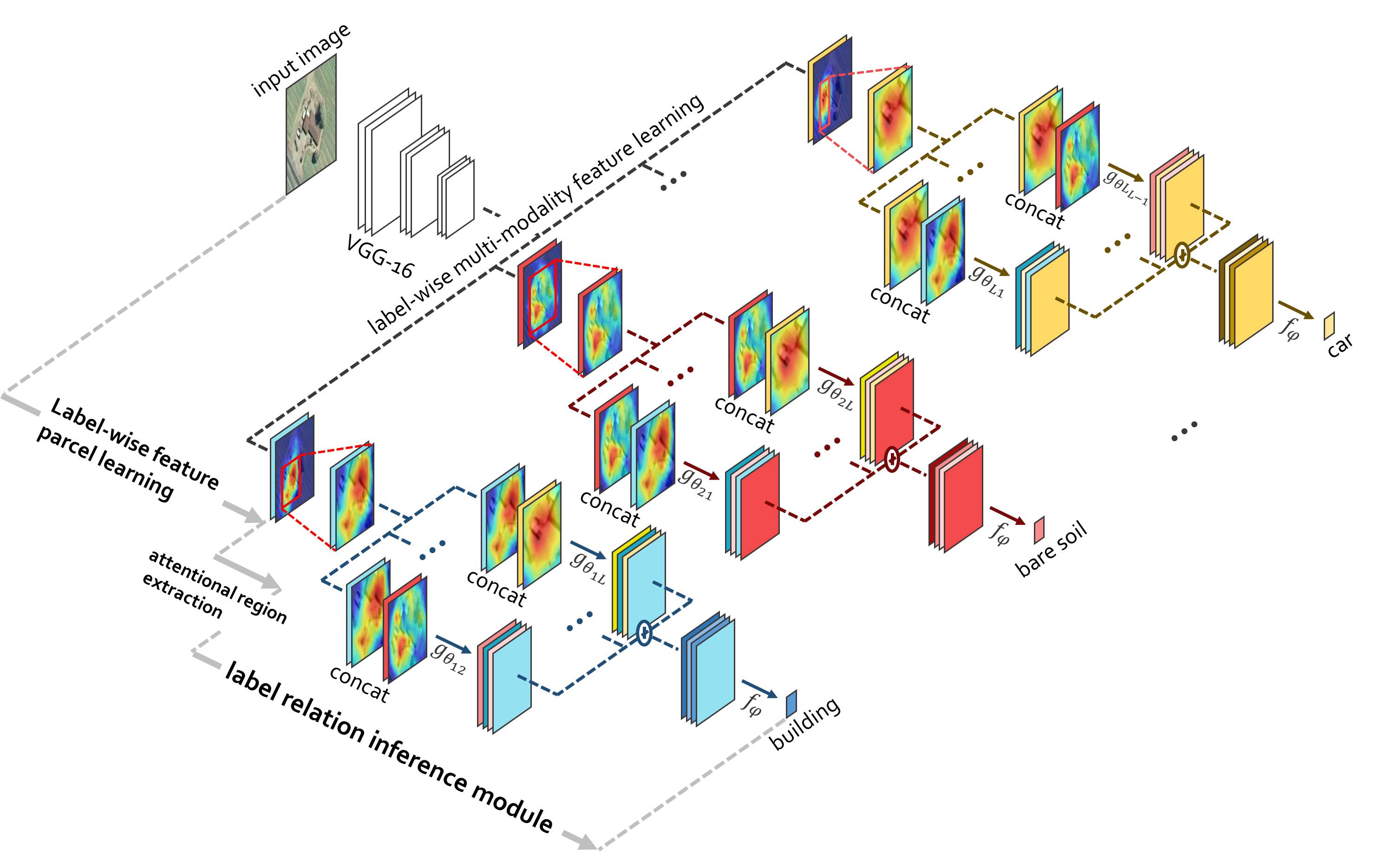}
    \caption{The architecture of the proposed attention-aware label relational reasoning network.}
    \label{fig:relation_net}
\end{figure*}
In order to explicitly model label relations, we propose a label relational inference network for multi-label aerial image classification. This work is inspired by recent successes of relation networks in visual question answering \cite{santoro2017simple}, object detection \cite{relationdet}, video classification \cite{wang2018appearance}, activity recognition in videos \cite{zhou2017temporal}, and semantic segmentation \cite{mou2019semantic}. A relation network is characterized by its inherent capability of inferring relations between an individual entity (e.g., a region in an image or a frame in a video) and all other entities (e.g., all regions in the image or all frames in the video). Besides, to increase the effectiveness of relational reasoning, we make use of a spatial transformer, which is often used to enhance the transformation invariance of deep neural networks~\cite{jaderberg2015spatial}, to reduce the impact of irrelevant semantic features.

More specifically, in this work, an innovative end-to-end multi-label aerial image classification network, termed as attention-aware label relational reasoning network, is proposed and characterized by its capabilities of localizing label-specific discriminative regions and explicitly modeling semantic label dependencies for the task. This paper's contributions are threefold.

\begin{itemize}
  \item We propose a novel multi-label aerial image classification network, attention-aware label relational reasoning network, which consists of three imperative components: a label-wise feature parcel learning module, an attentional region extraction module, and a label relational inference module. To our best knowledge, it is the first time that the idea of relation networks is employed to predict multiple object labels of aerial images, and experimental results demonstrate its effectiveness.
  
  \item We extract attentional regions from the label-wise feature parcels in a proposal-free fashion. Particularly, a learnable spatial transformer is employed to localize attentional regions, which are assumed to contain discriminative information, and then re-coordinate them into a given size. By doing so, attentional feature parcels can be yielded.
  
  \item To facilitate progress in the multi-label aerial image classification, we produce a new dataset, AID multi-label dataset, by relabeling images in the AID dataset \cite{xia2017aid}. In comparison with the UCM multi-label dataset \cite{chaudhuri2018multilabel}, the proposed dataset is more challenging due to diverse spatial resolutions of images, more scenes, and more samples. 
  
\end{itemize}

The remaining sections of this paper are organized as follows. Section \ref{sec:method} delineates three elemental modules of our proposed network, and Section \ref{sec:experiment} introduces experiments, where experimental setups are given and results are analyzed and discussed. Eventually, Section \ref{sec:conclusion} draws a conclusion of this paper.

\section{Methodology}
\label{sec:method}
\begin{figure*}[!t]
    \centering
    \includegraphics[width=0.8\textwidth]{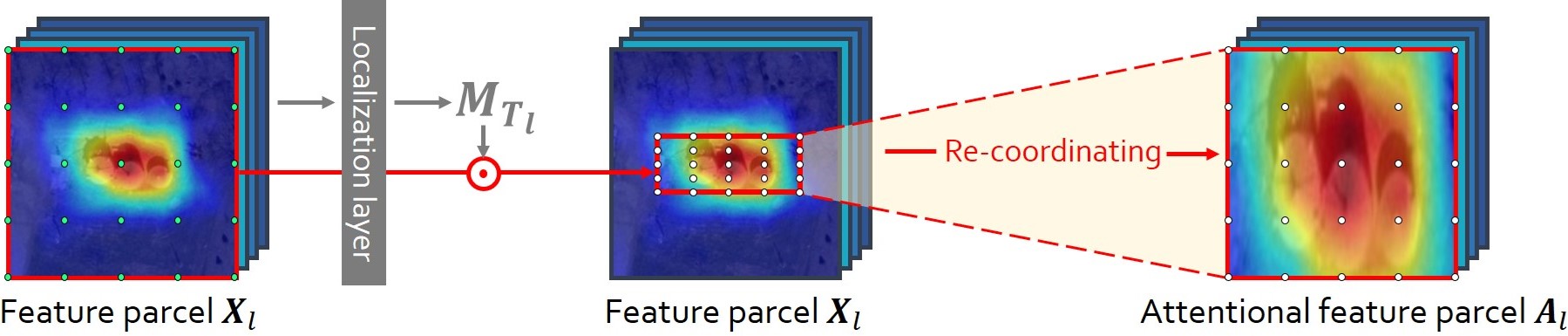}
    \caption{Illustration of the attentional region extraction module. Green dots in the left image indicate the feature parcel grid $G_{\bm{X}_l}$. White dots in the middle image represent the attentional feature parcel grid $G_{\bm{X}_l^{attn}}$, while those in the right image indicate re-coordinated $G_{\bm{X}_l^{attn}}$. Notably, the structure of re-coordinated $G_{\bm{X}_l^{attn}}$ is identical to that of $G_{\bm{X}_l}$, and values of pixels located at grid points in re-coordinated $G_{\bm{X}_l^{attn}}$ are obtained from those in $G_{\bm{X}_l^{attn}}$. For example, the pixel at the left top corner grid point in re-coordinated $G_{\bm{X}_l^{attn}}$ is assigned with the value of that at the left top corner of $G_{\bm{X}_l^{attn}}$.}
    \label{fig:are}
\end{figure*}
\subsection{Network Architecture}
As illustrated in Fig. \ref{fig:relation_net}, the proposed network comprises three components: a label-wise feature parcel learning module, an attentional region extraction module, and a label relational inference module. Let $L$ be the number of object labels and $l$ be the $l$-th label. The label-wise feature parcel learning module is designed to extract high-level feature maps $\bm{X}_l$ with $K$ channels, termed as \textit{feature parcel} (for more details refer to Section \ref{sec:p1}), for each label $l$. The attentional region extraction module is used to localize discriminative regions in each $\bm{X}_l$ and generate an attentional feature parcel $\bm{A}_l$, which is supposed to contain the most relevant semantics with respect to the label $l$. Finally, relations among $\bm{A}_l$ and all other label-wise attentional feature parcels are reasoned about by the label relational inference module for predicting the presence of the object $l$.
\par
Details of the proposed network are introduced in the remaining sections. 

\subsection{Label-wise Feature Parcel Learning}
\label{sec:p1}

The extraction of high-level features is crucial for visual recognition tasks, and many recent studies adopt CNNs owing to their remarkable performance in learning such features \cite{zhu2017deep, xia2019mining, zhang2016deep, srivastava2019understanding, wang2019igarss, qiu2019local, qiu2019fusingLetter}. Hence, we take a standard CNN as the backbone of the label-wise feature parcel learning module in our model. As shown in Fig.~\ref{fig:relation_net}, an aerial image is first fed into a CNN (e.g., VGG-16), which consists of only convolutional and max-pooling layers, for generating high-level feature maps. Subsequently, these features are encoded into $L$ feature parcels for each label $l$ via a label-wise multi-modality feature learning layer. To implement this layer, we first employ a convolutional layer with $KL$ filters, whose size is $1 \times 1$, to extract $KL$ feature maps. Afterwards, we divide these features into $L$ feature parcels, and each includes $K$ feature maps. That is to say, for each label, $K$ specific feature maps are learned, so-called \textit{feature parcel}, to extract discriminative semantics after the end-to-end training of the whole network. We denote the feature parcel for label $l$ as $\bm{X}_l$ in the following statements. 

In our experiment, we notice that $\bm{X}_l$ with a higher resolution is beneficial for the subsequent module to localize discriminative regions, as more spatial contextual cues are included. Accordingly, we discard the last max-pooling layer in VGG-16, leading to a spatial size of $14 \times 14$ for outputs. Weights are initialized with pre-trained VGG-16 on ImageNet but updated during the training phase.


\subsection{Attentional Region Extraction Module}
\label{sec:p2}
Although label-wise feature parcels can be directly applied to exploring label dependencies \cite{hua2019recurrently}, less informative regions (see blue areas in Fig. \ref{fig:are}) may bring noise and further reduce the effectiveness of these feature parcels. As shown in the left image of Fig. \ref{fig:are}, weakly activated regions indicate a loose relevance to the corresponding label, while highlighted regions suggest a strong region-label relevance. To diminish the influence of unrelated regions, we employ an attentional region extraction module to automatically extract discriminative regions from label-wise feature parcels.
\par
We localize and re-coordinate attentional regions from $\bm{X}_l$ with a learnable spatial transformer. Particularly, we sample a feature parcel $\bm{X}_l$ into a regular spatial grid $G_{\bm{X}_l}$ (cf. green dots in the left image of Fig.~\ref{fig:are}) according to the spatial resolution of $\bm{X}_l$ and regard pixels in $\bm{X}_l$ as points on the grid $G_{\bm{X}_l}$ with coordinates $(x_l, y_l)$. Similarly, we can define coordinates of a new grid, attentional region grid $G_{\bm{X}_l^{attn}}$ (see white dots in the middle image of Fig.~\ref{fig:are}), as $(x_l^{attn}, y_l^{attn})$, and the number of grid points along with the height and width is equivalent to that of $G_{\bm{X}_l}$. As demonstrated in \cite{jaderberg2015spatial} that $G_{\bm{X}_l^{attn}}$ can be learned by performing spatial transformation on $G_{\bm{X}_l}$, $(x_l^{attn}, y_l^{attn})$ can be calculated with the following equation:
\begin{equation}
\label{eq:st}
    \begin{bmatrix}
        x_l^{attn} \\
        y_l^{attn}
    \end{bmatrix} 
    = 
    \bm{M}_{T_l}
    \begin{bmatrix}
        x_l \\
        y_l \\
        1
    \end{bmatrix},
\end{equation}
where $\bm{M}_{T_l}$ is a learnable transformation matrix, and grid coordinates, $x_l$ and $y_l$, are normalized to $[-1, 1]$. Considering that this module is designed for localization, we only adopt scaling and translation in our case. Hence Eq. \ref{eq:st} can be rewritten as
\begin{equation}
\label{eq:st2}
    \begin{bmatrix}
        x_l^{attn} \\
        y_l^{attn}
    \end{bmatrix} 
    = 
    \begin{bmatrix}
        s_{x_l} & 0 & t_{x_l} \\
        0 & s_{y_l} & t_{y_l} \\
    \end{bmatrix}
    \begin{bmatrix}
        x_l \\
        y_l \\
        1
    \end{bmatrix}
    ,
\end{equation}
where $s_{x_l}$ and $s_{y_l}$ indicate scaling factors along x- and y-axis, respectively, and $t_{x_l}$ and $t_{y_l}$ represent how feature maps should be translated along both axes. Notably, since different objects distribute variously in aerial images, $\bm{M}_{T_l}$ is learned for each object label $l$ individually. In other words, extracted attentional regions are label-specific and capable of improving the effectiveness of label-wise features. 
\par
As to the implementation of this module, we first vectorize $\bm{X}_l$ with a flatten function and then employ a localization layer (e.g., a fully connected layer) to estimate elements in $\bm{M}_{T_l}$ from the vectorized $\bm{X}_l$. Afterwards, attentional region grid coordinates $(x_l^{attn}, y_l^{attn})$ can be learned from $(x_l, y_l)$ with Eq.~\ref{eq:st2}, and values of pixels at $(x_l^{attn}, y_l^{attn})$ is able to be obtained from neighboring pixels by bilinear interpolation. Finally, the attentional region grid $G_{\bm{X}_l^{attn}}$ is re-coordinated to a regular spatial grid, which shares an identical structure with $G_{\bm{X}_l}$, for yielding the final attentional feature parcel $\bm{A}_l$.


\subsection{Label Relational Inference Module}
\label{sec:p3}
\begin{figure*}[!t]
    \centering
    \includegraphics[width=.9\textwidth]{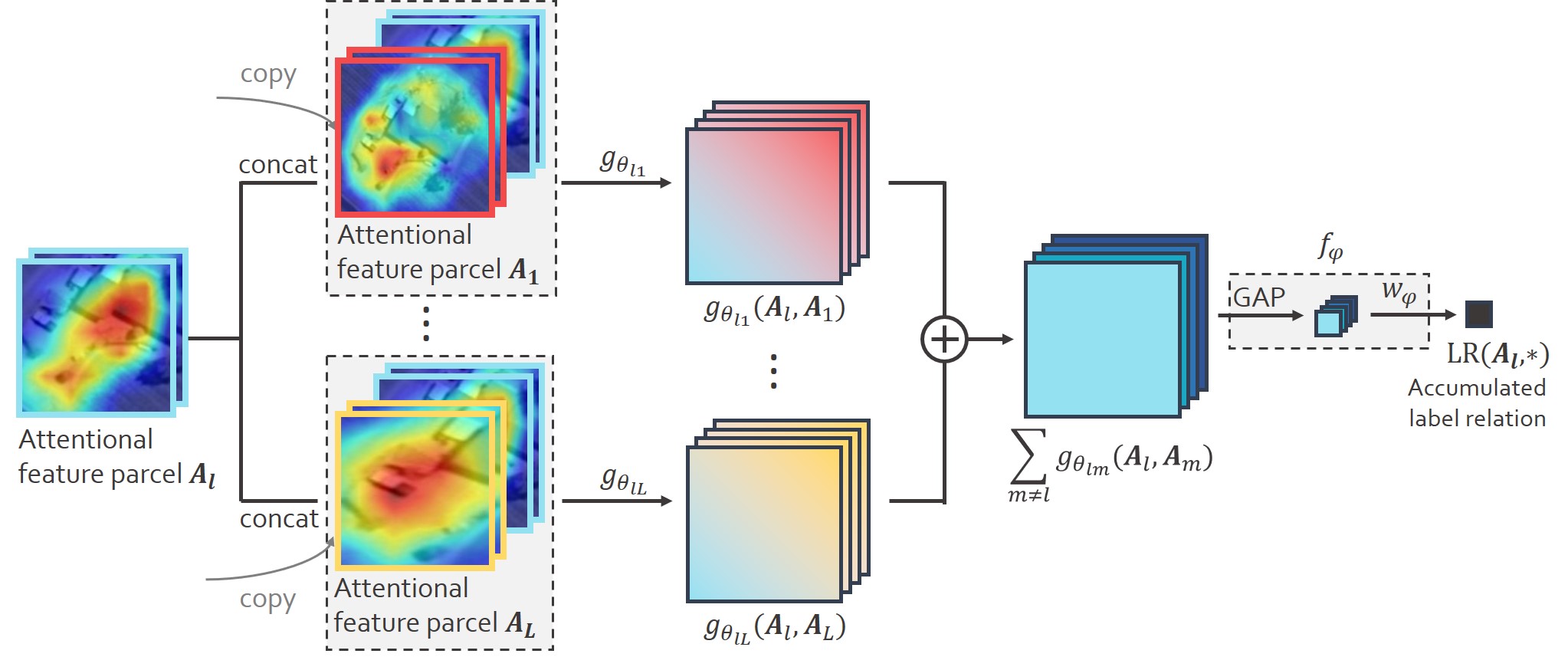}
    \caption{Illustration of the label relation module.}
    \label{fig:relation_module}
\end{figure*}

Being the core of our model, the label relational inference module is designed to fully exploit label interrelations for inferring existences of all labels. Before diving into this module, we define the pairwise label relation as a composite function with the following equation:
\begin{equation}
\label{eq:relation}
    {\rm LR}(\bm{A}_{l}, \bm{A}_{m}) = f_\phi(g_{\theta_{lm}}(\bm{A}_{l}, \bm{A}_{m})),
\end{equation}
where the input is a pair of attentional feature parcels, $\bm{A}_l$ and $\bm{A}_m$, and $l$ and $m$ range from $1$ to $L$. The functions $g_{\theta_{lm}}$ and $f_\phi$ are used to reason about the pairwise relation between label $l$ and $m$. More specifically, the role of $g_{\theta_{lm}}$ is to reason about whether there exist relations between the two objects and how they are related. In previous works~\cite{santoro2017simple, zhou2017temporal}, a multilayer perceptron (MLP) is commonly employed as $g_{\theta_{lm}}$ for its simplicity. However, spatial contextual semantics are not taken into account in this way. To address such issue, here, we make use of $1 \times 1$ convolution instead of an MLP to explore spatial information. Furthermore, $f_\phi$ is applied to encode the output of $g_{\theta_{lm}}$ into the final pairwise label relation ${\rm LR}(\bm{A}_{l}, \bm{A}_{m})$. In our case, $f_\phi$ consists of a global average pooling layer and an MLP, which finally yields the relation between label $l$ and $m$.

Following the motivation of our work, we infer each label by accumulating all related pairwise label relations, and the accumulated label relation for object label $l$ is defined as: 
\begin{equation}
\label{eq:relation2}
    {\rm LR}(\bm{A}_{l}, *) = f_\phi(\sum_{m \neq l}g_{\theta_{lm}}(\bm{A}_{l}, \bm{A}_{m})),
\end{equation}
where $*$ represents all attentional feature parcels except $\bm{A}_l$. Based on this formula, we implement the label relational inference module with the following steps (taking the prediction of label $l$ as an example): 1) $\bm{A}_l$ and every other attentional feature parcel are concatenated and fed into a $1\times 1$ convolutional layer, respectively. 2) Afterwards, a global average pooling layer is employed to transform $g_{\theta_{lm}}(\bm{A}_{l}, \bm{A}_{m})$ into vectors, which are then element-wise added. 3) Finally, the output is fed into an MLP layer with trainable parameters $\phi$ to produce the accumulated label relation ${\rm LR}(\bm{A}_{l}, *)$. Note that $g_{\theta_{lm}}$ is a learnable unit, which models pairwise relations using convolutions. Through the end-to-end training, it could be expected to learn data-driven label relations. Experiments in Section \ref{sec:out_ucm} and Section \ref{sec:out_aid} have verified that learned label relations are in line with prior knowledge. Since we expect the model to predict probabilities, an activation function $\sigma$ is utilized to restrict each output digit to $[0,1]$. For label $l$, a digit approaching $1$ implies a high probability of its presence, while one closing $0$ suggests the absence. Fig.~\ref{fig:relation_module} presents an visual illustration of the label relational inference module. 

Compared to other multi-label classification methods, our model has three benefits:
\begin{enumerate}
  \setlength{\itemsep}{0pt}
  \setlength{\parsep}{0pt}
  \setlength{\parskip}{0pt}
    \item The module can inherently reason about label relations as indicated by Eq.~\ref{eq:relation} and requires no particular prior knowledge about relations among all objects. That is to say, our network does not need to learn \textit{how to compute label relations} and \textit{which object relations should be considered}. All relations are automatically learned through a data-driven way and proven to meet the reality in our experiments.
    \item The learning effectiveness is independent of long short-term memory, leading to increased robustness. This is because, in Eq.~\ref{eq:relation2}, accumulated label relations are calculated with a summation function instead of a chain architecture, e.g., an LSTM.
    \item The function $g_{\theta_{lm}}$ is learned for each object label pair $l$ and $m$ separately, which suggests that pairwise label relations are encoded in a specific way. Besides, our implementation of $g_{\theta_{lm}}$ can extend the applicability of relational reasoning compared to using an MLP.
\end{enumerate}

Since \cite{hua2019recurrently} shares the same design philosophy that modeling label relations is crucial, here we emphasize two differences between our network and \cite{hua2019recurrently}: 1) the proposed network learns to extract discriminative regions as label-wise features for modeling label relations (cf. Section \ref{sec:p2}) instead of directly using entire feature maps as in \cite{hua2019recurrently}; 2) the proposed label relation inference module encodes label relations explicitly with composite functions, while in \cite{hua2019recurrently}, label relations are modeled implicitly via an RNN whose effectiveness depends heavily on the learning effect of long-term memorization. Quantitative comparisons between these two approaches are shown in the following section.

\section{Experiments and Discussion}
\label{sec:experiment}

In this section, we conduct experiments on the UCM \cite{chaudhuri2018multilabel} and proposed AID multi-label dataset for evaluating our model. Specifically, Section \ref{sec:data} presents a description of these two datasets. Afterwards, we introduce training strategies and thoroughly discuss experimental results in the subsequent subsections.

\subsection{Dataset Introduction}
\label{sec:data}
\begin{figure*}[!t]
\captionsetup[subfigure]{captionskip=0.1em}
\centering
\subfloat[]{\includegraphics[width=0.13\textwidth]{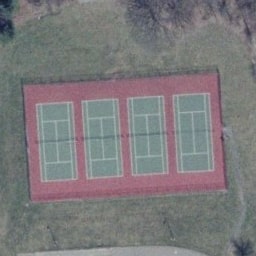}
\label{fig:s1}}
\hfil
\subfloat[]{\includegraphics[width=0.13\textwidth]{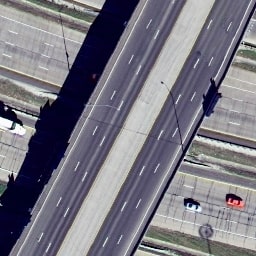}
\label{fig:s2}}
\hfil
\subfloat[]{\includegraphics[width=0.13\textwidth]{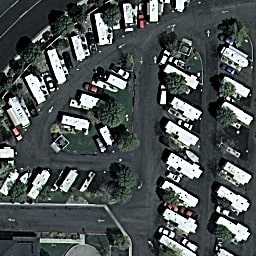}
\label{fig:s3}}
\hfil
\subfloat[]{\includegraphics[width=0.13\textwidth]{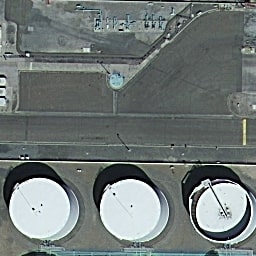}
\label{fig:s4}}
\hfil
\subfloat[]{\includegraphics[width=0.13\textwidth]{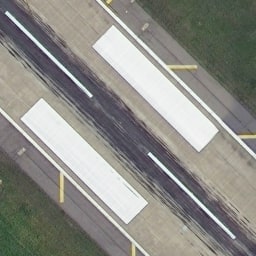}
\label{fig:s5}}
\hfil
\subfloat[]{\includegraphics[width=0.13\textwidth]{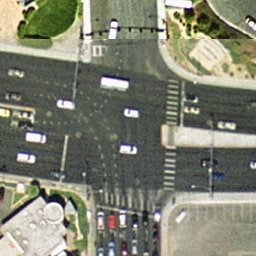}
\label{fig:s6}}
\hfil
\subfloat[]{\includegraphics[width=0.13\textwidth]{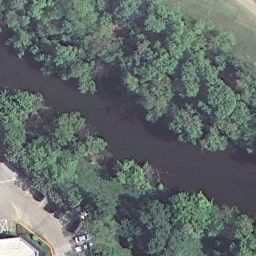}
\label{fig:s7}}
\vspace{-0.9em}
\subfloat[]{\includegraphics[width=0.13\textwidth]{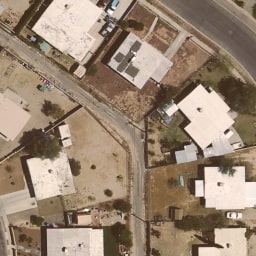}
\label{fig:s8}}
\hfil
\subfloat[]{\includegraphics[width=0.13\textwidth]{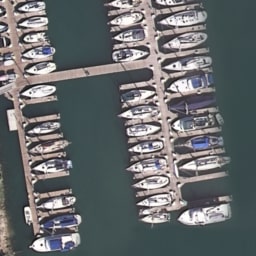}
\label{fig:s9}}
\hfil
\subfloat[]{\includegraphics[width=0.13\textwidth]{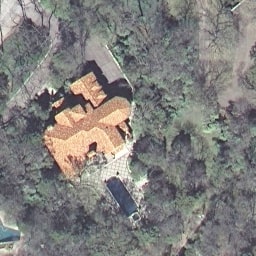}
\label{fig:s10}}
\hfil
\subfloat[]{\includegraphics[width=0.13\textwidth]{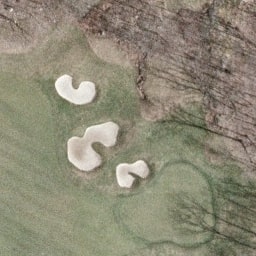}
\label{fig:s11}}
\hfil
\subfloat[]{\includegraphics[width=0.13\textwidth]{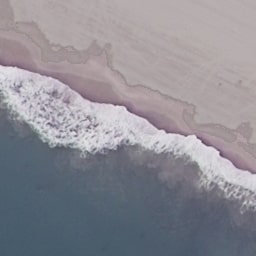}
\label{fig:s12}}
\hfil
\subfloat[]{\includegraphics[width=0.13\textwidth]{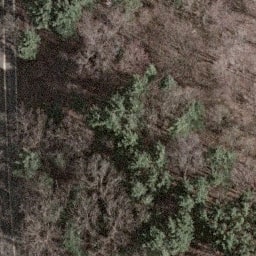}
\label{fig:s13}}
\hfil
\subfloat[]{\includegraphics[width=0.13\textwidth]{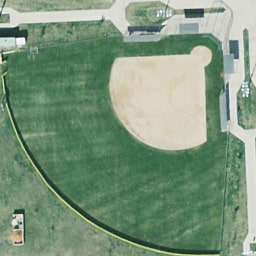}
\label{fig:s14}}
\vspace{-0.9em}
\subfloat[]{\includegraphics[width=0.13\textwidth]{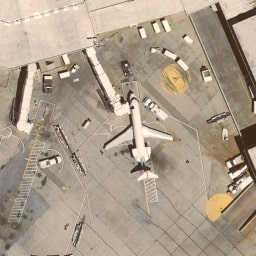}
\label{fig:s15}}
\hfil
\subfloat[]{\includegraphics[width=0.13\textwidth]{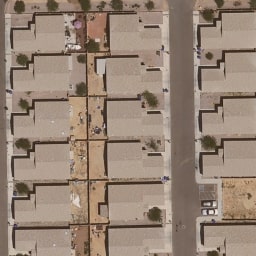}
\label{fig:s16}}
\hfil
\subfloat[]{\includegraphics[width=0.13\textwidth]{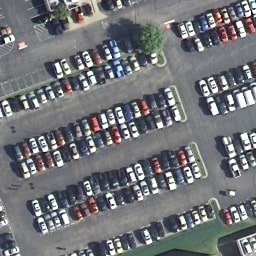}
\label{fig:s17}}
\hfil
\subfloat[]{\includegraphics[width=0.13\textwidth]{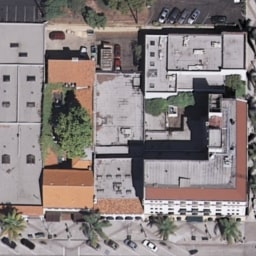}
\label{fig:s18}}
\hfil
\subfloat[]{\includegraphics[width=0.13\textwidth]{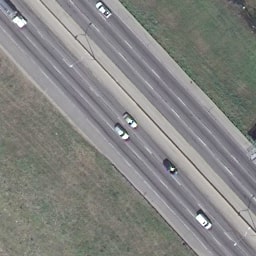}
\label{fig:s19}}
\hfil
\subfloat[]{\includegraphics[width=0.13\textwidth]{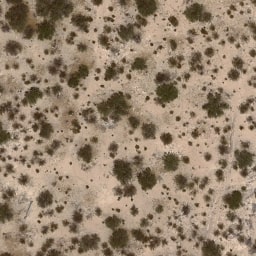}
\label{fig:s20}}
\hfil
\subfloat[]{\includegraphics[width=0.13\textwidth]{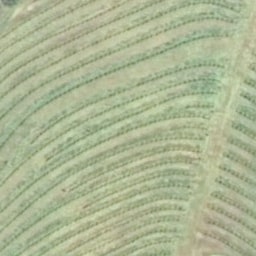}
\label{fig:s21}}
\caption{Samples of various scene categories in the UCM multi-label dataset as well as associated \textsl{object} labels. The spatial resolution of each image is one foot, and the size is $256 \times 256$ pixels. Scene and \textsl{object} labels of each sample are as follows: (a) Tennis court: \textsl{tree}, \textsl{grass}, \textsl{court}, and \textsl{bare soil}. (b) Overpass: \textsl{pavement}, \textsl{bare soil}, and \textsl{car}. (c) Mobile home park: \textsl{pavement}, \textsl{grass}, \textsl{bare soil},  \textsl{tree}, \textsl{mobile home}, and \textsl{car}. (d) Storage tank: \textsl{tank}, \textsl{pavement}, and \textsl{bare soil}. (e) Runway: \textsl{pavement} and \textsl{grass}. (f) Intersection: \textsl{car},  \textsl{tree}, \textsl{pavement}, \textsl{grass}, and \textsl{building}. (g) River: \textsl{water}, \textsl{tree}, and \textsl{grass}. (h) Medium residential: \textsl{pavement}, \textsl{grass}, \textsl{car}, \textsl{tree}, and \textsl{building}. (i) Harbor: \textsl{ship}, \textsl{water}, and \textsl{dock}. (j) Sparse residential: \textsl{car}, \textsl{tree}, \textsl{grass}, \textsl{pavement}, \textsl{building}, and \textsl{bare soil}. (k) Golf course: \textsl{sand}, \textsl{pavement},\textsl{tree}, and \textsl{grass}. (l) Beach: \textsl{sea} and \textsl{sand}. (m) Forest: \textsl{tree}, \textsl{grass}, and \textsl{building}. (n) Baseball diamond: \textsl{pavement}, \textsl{grass}, \textsl{building}, and \textsl{bare soil}. (o) Airplane: \textsl{airplane}, \textsl{car}, \textsl{bare soil}, \textsl{grass} and \textsl{pavement}. (p) Dense residential: \textsl{tree}, \textsl{building}, \textsl{pavement}, \textsl{grass}, and \textsl{car}. (q) Parking lot: \textsl{pavement}, \textsl{grass}, and \textsl{car}. (r) building: \textsl{pavement}, \textsl{car}, and \textsl{building}. (s) Free way: \textsl{tree}, \textsl{car}, \textsl{pavement}, \textsl{grass}, and \textsl{bare soil}. (t) Chaparral: \textsl{chaparral} and \textsl{bare soil}. (u) Agricultural: \textsl{tree} and \textsl{field}.}
\label{fig:samp1}
\end{figure*}
\subsubsection{UCM multi-label dataset}

UCM multi-label dataset \cite{chaudhuri2018multilabel} is reproduced by assigning all aerial images collected in UCM dataset \cite{yang2010bag} with newly defined object labels. The number of all candidate object labels is 17: building, sand, dock, court, tree, sea, bare soil, mobile home, ship, field, tank, water, grass, pavement, chaparral, and car. It is worth noting that labels, such as tank, airplane, and building, exist in both \cite{yang2010bag} and \cite{chaudhuri2018multilabel} while at different levels. In \cite{yang2010bag}, such terms are considered as scene-level labels due to the fact that related images can be characterized and depicted by them, while in \cite{chaudhuri2018multilabel}, they mean objects that may present in aerial images.

As to properties of images in this dataset, the spatial resolution of each sample is one foot, and the size is $256 \times 256$ pixels. All images are manually cropped from aerial imagery contributed by the National Map of the U.S. Geological Survey (USGS), and there are 2100 images in total. For each object category, the number of images is listed in Table \ref{tab:u}. Besides, 80\% of image samples per scene class are selected to train our model, and the other 20\% of images are used to test our model. Numbers of images assigned to training and test sets with respect to all object labels are available in Table \ref{tab:u} as well. Some visual examples are shown in Fig. \ref{fig:samp1}.

\begin{table}[!t]
\renewcommand{\arraystretch}{1.1}
\caption{The number of images for different object categories in the UCM multi-label dataset.}
\label{tab:u}
\centering
\begin{tabular}{c|cccc}
\Xhline{3\arrayrulewidth}
Category No. & Category Name & Training & Test & Total\\
\hline
\hline
1 & bare soil & 577 & 141 & 718 \\
2 & airplane & 80 & 20 & 100 \\
3 & building & 555 & 136 & 691 \\
4 & car & 722 & 164 & 886 \\
5 & chaparral & 82 & 33 & 115 \\
6 & court & 84 & 21 & 105 \\
7 & dock & 80 & 20 & 100 \\
8 & field & 79 & 25 & 104 \\
9 & grass & 804 & 171 & 975 \\
10 & mobile home & 82 & 20 & 102 \\
11 & pavement & 1047 & 253 & 1300 \\
12 & sand & 218 & 76 & 294 \\
13 & sea & 80 & 20 & 100 \\
14 & ship & 80 & 22 & 102 \\
15 & tank & 80 & 20 & 100 \\
16 & tree & 801 & 208 & 1009 \\
17 & water & 161 & 42 & 203 \\
\hline
\hline
- & All & 1680 & 420 & 2100 \\
\Xhline{3\arrayrulewidth}
\end{tabular}
\end{table}

\subsubsection{AID multi-label dataset}
\begin{figure*}[!t]
\captionsetup[subfigure]{captionskip=0.1em}
\centering
\subfloat[]{\includegraphics[width=0.137\textwidth]{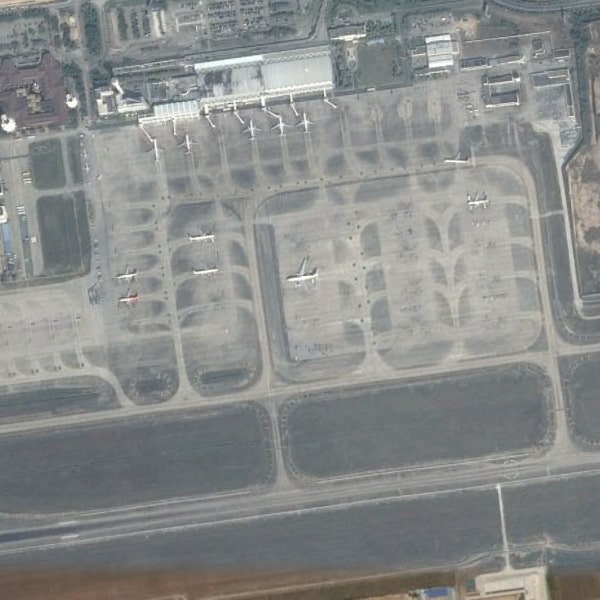}
\label{fig:aids1}}
\hfil
\subfloat[]{\includegraphics[width=0.137\textwidth]{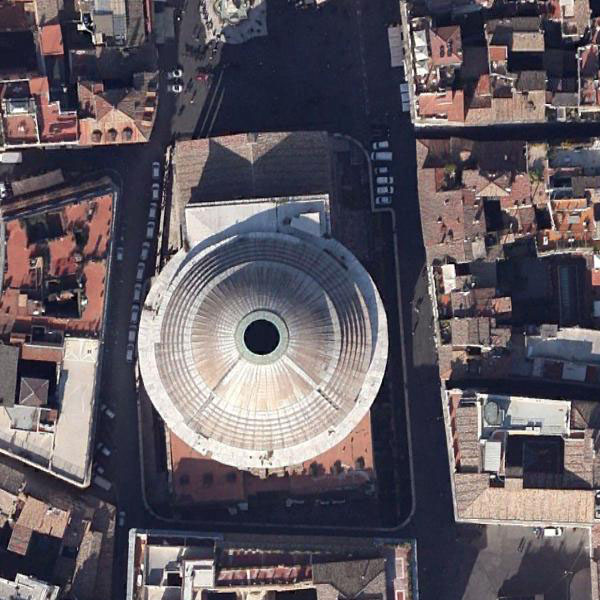}
\label{fig:aids2}}
\hfil
\subfloat[]{\includegraphics[width=0.137\textwidth]{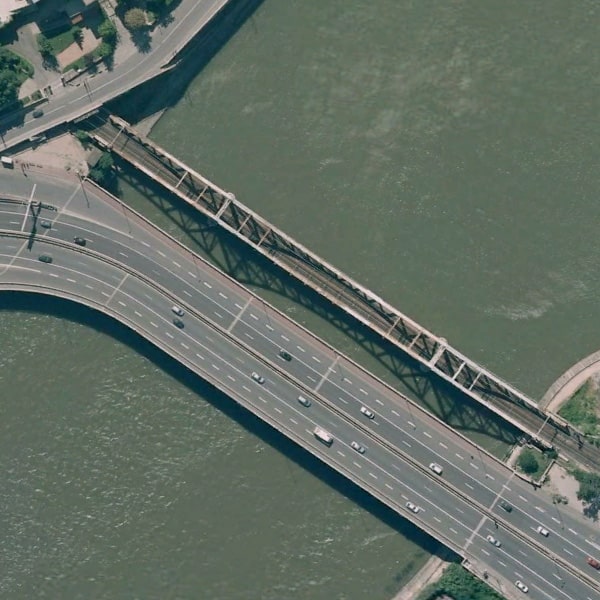}
\label{fig:aids5}}
\hfil
\subfloat[]{\includegraphics[width=0.137\textwidth]{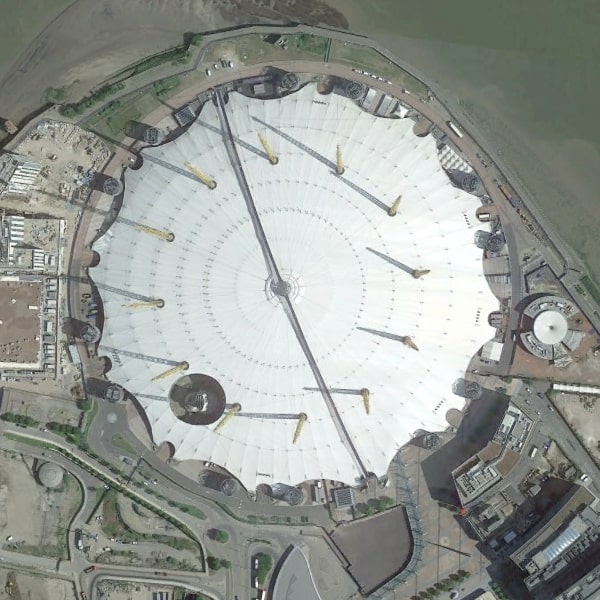}
\label{fig:aids6}}
\hfil
\subfloat[]{\includegraphics[width=0.137\textwidth]{samples/bareland_66.jpg}
\label{fig:aids7}}
\hfil
\subfloat[]{\includegraphics[width=0.137\textwidth]{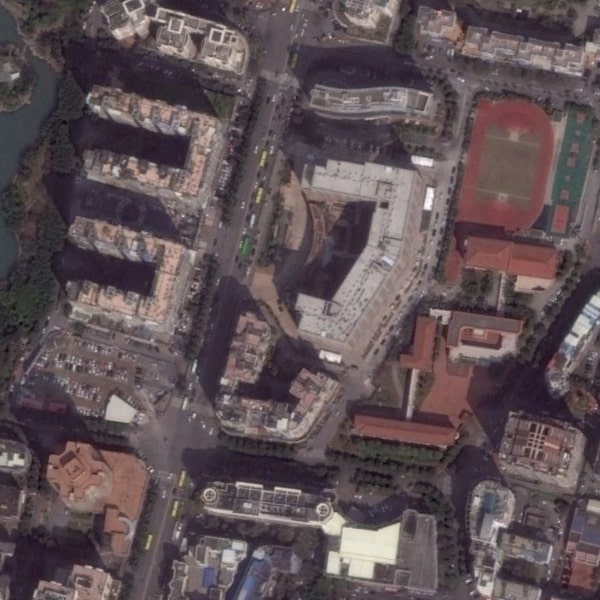}
\label{fig:aids8}}
\vspace{-0.9em}
\subfloat[]{\includegraphics[width=0.137\textwidth]{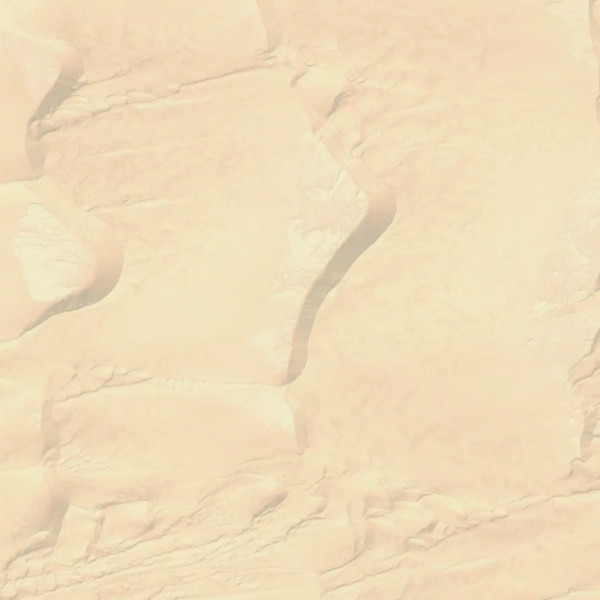}
\label{fig:aids10}}
\hfil
\subfloat[]{\includegraphics[width=0.137\textwidth]{samples/forest_17.jpg}
\label{fig:aids12}}
\hfil
\subfloat[]{\includegraphics[width=0.137\textwidth]{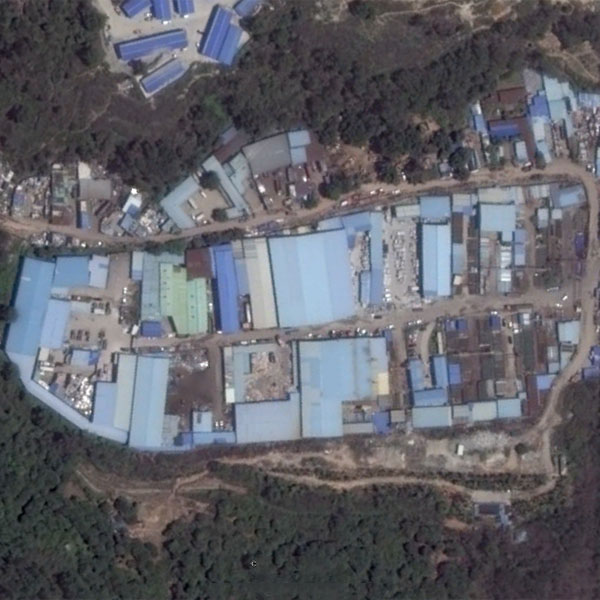}
\label{fig:aids13}}
\hfil
\subfloat[]{\includegraphics[width=0.137\textwidth]{samples/meadow_52.jpg}
\label{fig:aids14}}
\hfil
\subfloat[]{\includegraphics[width=0.137\textwidth]{samples/mountain_10.jpg}
\label{fig:aids16}}
\hfil
\subfloat[]{\includegraphics[width=0.137\textwidth]{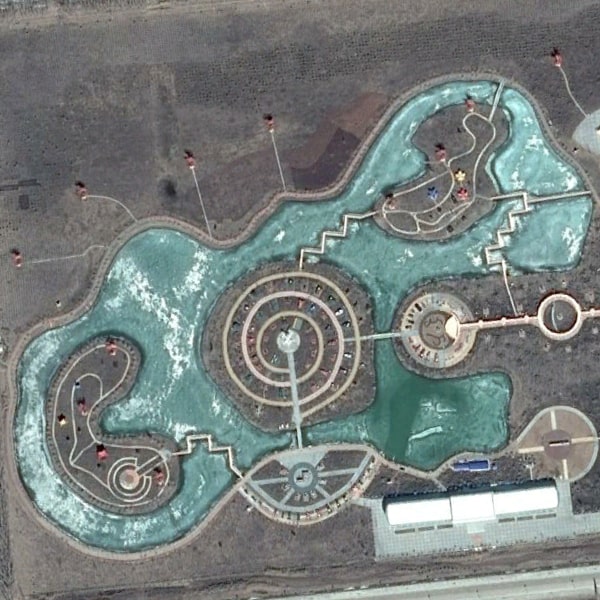}
\label{fig:aids17}}
\vspace{-0.9em}
\subfloat[]{\includegraphics[width=0.137\textwidth]{samples/playground_14.jpg}
\label{fig:aids19}}
\hfil
\subfloat[]{\includegraphics[width=0.137\textwidth]{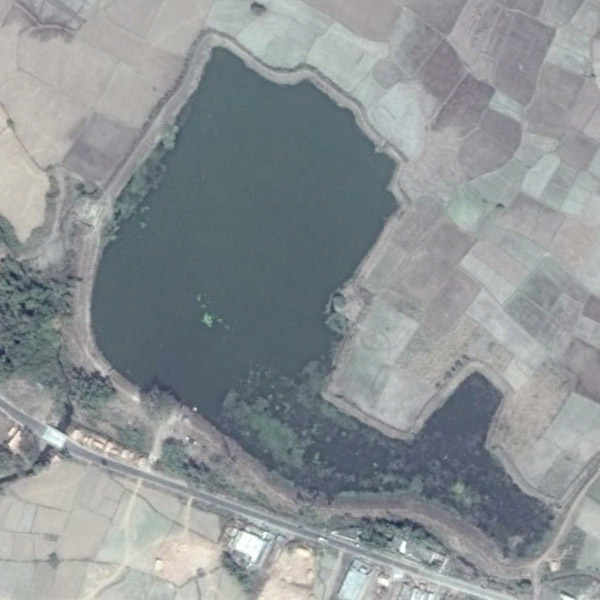}
\label{fig:aids20}}
\hfil
\subfloat[]{\includegraphics[width=0.137\textwidth]{samples/port_13.jpg}
\label{fig:aids21}}
\hfil
\subfloat[]{\includegraphics[width=0.137\textwidth]{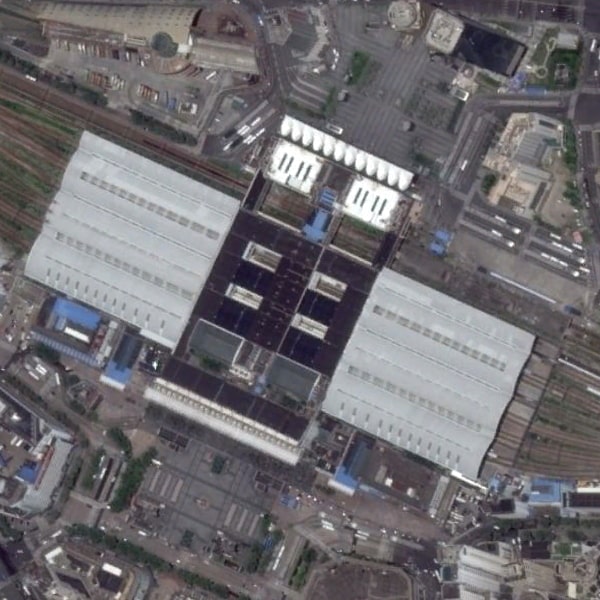}
\label{fig:aids22}}
\hfil
\subfloat[]{\includegraphics[width=0.137\textwidth]{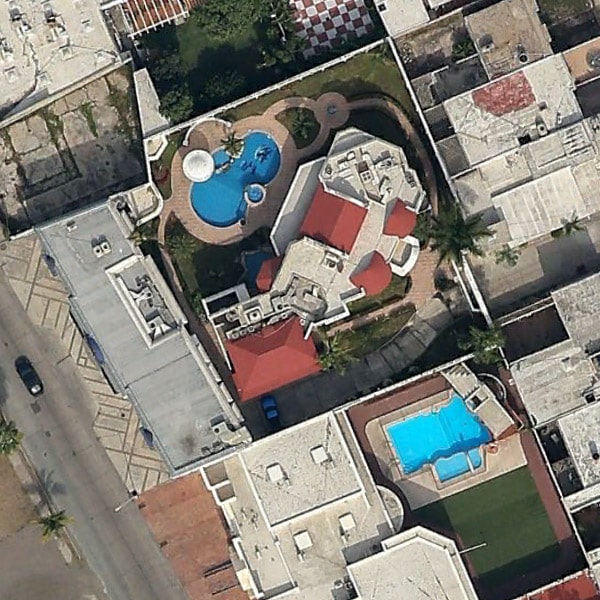}
\label{fig:aids23}}
\hfil
\subfloat[]{\includegraphics[width=0.137\textwidth]{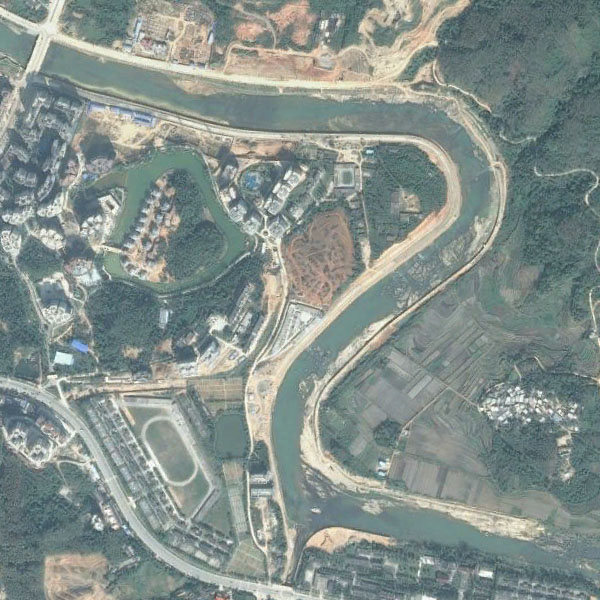}
\label{fig:aids24}}
\vspace{-0.9em}
\subfloat[]{\includegraphics[width=0.137\textwidth]{samples/school_7.jpg}
\label{fig:aids25}}
\hfil
\subfloat[]{\includegraphics[width=0.137\textwidth]{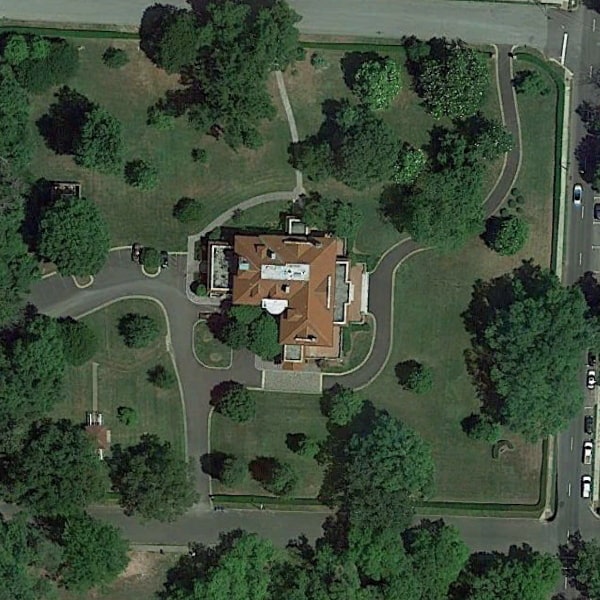}
\label{fig:aids26}}
\hfil
\subfloat[]{\includegraphics[width=0.137\textwidth]{samples/square_38.jpg}
\label{fig:aids27}}
\hfil
\subfloat[]{\includegraphics[width=0.137\textwidth]{samples/stadium_45.jpg}
\label{fig:aids28}}
\hfil
\subfloat[]{\includegraphics[width=0.137\textwidth]{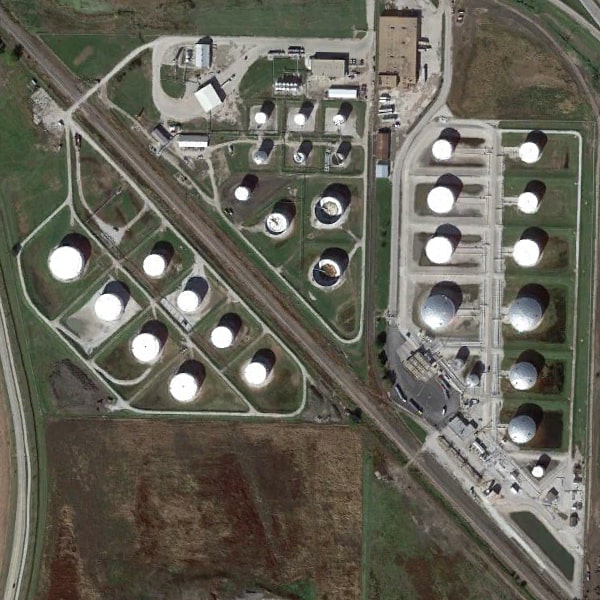}
\label{fig:s21s29}}
\hfil
\subfloat[]{\includegraphics[width=0.137\textwidth]{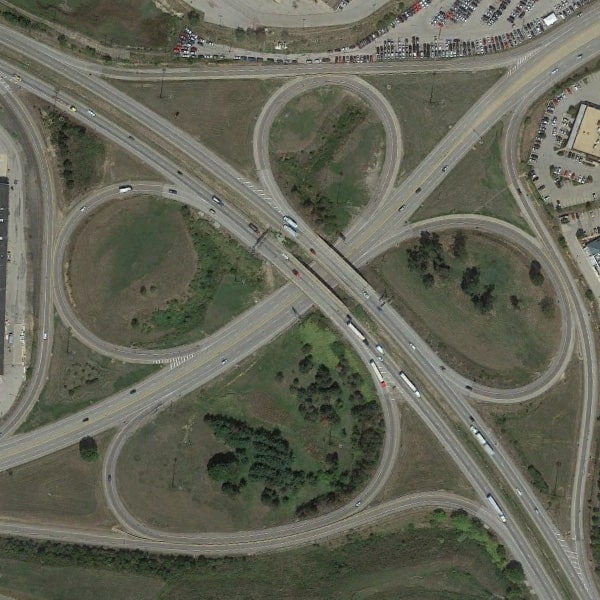}
\label{fig:s21s30}}

\caption{Samples of various scene categories in the AID multi-label dataset and their associated \textsl{object} labels. The spatial resolution of each image varies from 0.5 to 8 m/pixel, and the size is $600 \times 600$ pixels. Here are scene and \textsl{object} labels of selected samples: (a) Airport: \textsl{car}, \textsl{building}, \textsl{tank}, \textsl{tree}, \textsl{airplane}, \textsl{grass}, \textsl{pavement}, and \textsl{bare soil}. (b) Church: \textsl{pavement}, \textsl{car}, and \textsl{building}. (c) Bridge: \textsl{building}, \textsl{car}, \textsl{grass}, \textsl{pavement}, \textsl{tree} and \textsl{water}. (d) Center: \textsl{grass},  \textsl{building}, \textsl{tree}, \textsl{car}, \textsl{bare soil}, and \textsl{pavement}. (e) Bare land: \textsl{bare soil}, \textsl{building}, \textsl{pavement}, and \textsl{water}. (f) Commercial: \textsl{building}, \textsl{car}, \textsl{court}, \textsl{grass}, \textsl{pavement}, \textsl{tree}, and \textsl{water}. (g) Desert: \textsl{sand}. (h) Forest: \textsl{bare soil} and \textsl{tree}. (i) Industrial:  \textsl{pavement}, \textsl{grass}, \textsl{car}, \textsl{bare soil}, and \textsl{building}. (j) Meadow: \textsl{pavement} and \textsl{grass}. (k) Mountain: \textsl{tree} and \textsl{grass}. (l) Park: \textsl{bare soil}, \textsl{building}, \textsl{court}, \textsl{grass}, \textsl{pavement}, \textsl{tree}, and \textsl{water}. (m) Playground: \textsl{car}, \textsl{grass}, and \textsl{pavement}. (n) Pond: \textsl{building}, \textsl{field}, \textsl{grass}, \textsl{pavement}, \textsl{tree}, and \textsl{water}. (o) Port: \textsl{ship}, \textsl{sea}, \textsl{car},  \textsl{grass}, \textsl{pavement}, \textsl{tree}, \textsl{building}, and \textsl{dock}. (p) Railway: \textsl{tree}, \textsl{car}, \textsl{pavement}, \textsl{building}, and \textsl{grass}. (q) Resort: \textsl{pavement}, \textsl{building}, \textsl{car}, \textsl{tree}, \textsl{field}, \textsl{bare soil}, and \textsl{water}. (r) River: \textsl{car}, \textsl{building}, \textsl{bare soil}, \textsl{dock}, \textsl{water}, \textsl{grass}, \textsl{pavement}, \textsl{tree}, \textsl{ship}, and \textsl{field}. (s) School: \textsl{pavement}, \textsl{tank}, \textsl{grass}, \textsl{court}, \textsl{building}, and \textsl{car}. (t) Sparse residential: \textsl{pavement}, \textsl{car}, \textsl{building}, \textsl{tree}, and \textsl{grass}. (u) Square: \textsl{tree}, \textsl{car}, \textsl{court}, \textsl{pavement}, \textsl{grass}, and \textsl{building}. (v) Stadium: \textsl{car}, \textsl{pavement}, \textsl{tree}, \textsl{court}, \textsl{grass}, \textsl{building}, and \textsl{bare soil}. (w) Storage tanks: \textsl{tank}, \textsl{tree}, \textsl{car}, \textsl{grass}, \textsl{pavement}, \textsl{building}, and \textsl{bare soil}. (x) Viaduct: \textsl{pavement}, \textsl{car}, \textsl{bare soil}, \textsl{tree}, \textsl{grass}, and \textsl{building}.}
\label{fig:samp2}
\end{figure*}
In order to further evaluate our network and meanwhile promote progress in the area of multi-class classification of high-resolution aerial images, we produce a new dataset, named AID multi-label dataset, based on the widely used AID scene classification dataset \cite{xia2017aid}. The AID dataset consists of 10000 high-resolution aerial images collected from worldwide Google Earth imagery, including scenes from China, the United States, England, France, Italy, Japan, and Germany. In contrast to the UCM dataset, spatial resolutions of images in the AID dataset vary from 0.5 m/pixel to 8 m/pixel, and the size of each aerial image is $600 \times 600$ pixels. Besides, the number of images in each scene category ranges from 220 to 420. Overall, the AID dataset is more challenging compared to the UCM dataset.

Here, we manually relabel some images in the AID dataset. With extensive human visual inspections, 3000 aerial images from 30 scenes in the AID dataset are selected and assigned with multiple object labels, and the distribution of samples in each category is shown in Table \ref{tab:a}. Besides, 80\% of all images are taken as training samples, while the rest is used for testing our model. Several example images are shown in Fig. \ref{fig:samp2}.

\subsection{Training Details}
\label{sec:train}

\begin{table}[!t]
\renewcommand{\arraystretch}{1.1}
\caption{The number of images for different object categories in the AID multi-label dataset.}
\label{tab:a}
\centering
\begin{tabular}{c|cccc}
\Xhline{3\arrayrulewidth}
Category No. & Category Name & Training & Test & Total \\
\hline
\hline
1 & bare soil & 1171 & 304 & 1475\\
2 & airplane & 79 & 20 & 99 \\
3 & building & 1744 & 417 & 2161 \\
4 & car & 1617 & 409 & 2026 \\
5 & chaparral & 75 & 37 & 112 \\
6 & court & 269 & 75 & 344 \\
7 & dock & 221 & 50 & 271 \\
8 & field & 175 & 39 & 214 \\
9 & grass & 1829 & 466 & 2295 \\
10 & mobile home & 1 & 1 & 2 \\
11 & pavement & 1870 & 458 & 2328\\
12 & sand & 207 & 52 & 259 \\
13 & sea & 177 & 44 & 221 \\
14 & ship & 237 & 47 & 284 \\
15 & tank & 87 & 21 & 108 \\
16 & tree & 1923 & 483 & 2406 \\
17 & water & 674 & 178 & 852 \\
\hline
\hline
- & All & 2400 & 600 & 3000 \\
\Xhline{3\arrayrulewidth}
\end{tabular}
\end{table}

As to the initialization of our network, different modules are done in different ways. For the label-wise feature parcel learning module, we initialize the backbone and weights in other convolutional layers with a pre-trained ImageNet \cite{imagenet_cvpr09} model and a Glorot uniform initializer, respectively. Regarding the attentional region extraction module, we initialize the transformation matrix in Eq. \ref{eq:st} as an identical transformation,
\begin{equation}
\label{eq:m}
    \bm{M}_{T_l}= 
    \begin{bmatrix}
        1 & 0 & 0 \\
        0 & 1 & 0 \\
    \end{bmatrix}
    .
\end{equation}
In the label relational inference module, weights in both $f_\phi$ and $g_{\theta_{lm}}$ are initialized with a Glorot uniform initializer and updated during the training phase. Notably, the entire network is trained in an end-to-end manner, and weights in the backbone are fine-tuned as well.
\par
In our case, multiple labels are encoded into multi-hot binary sequences instead of one-hot vectors widely used in single-label classification tasks. The length of such multi-hot binary sequence is identical to the number of total object categories, i.e., 17 in our case, and as to each digit, 0 suggests an absent object, while 1 indicates the presence of its corresponding object label. Accordingly, we define the network loss as the binary cross-entropy. Besides, Adam with Nesterov momentum \cite{nadam2}, which shows faster convergence than stochastic gradient descent (SGD) for our task, are selected and its parameters are set as recommended \cite{nadam2}: $\epsilon = 1e-08$, $\beta_1 = 0.9$, and $\beta_2 = 0.999$. The learning rate is initially defined as $1e-04$ and decayed by a factor of 10 if the validation loss fails to decrease. Notably, we randomly select 10\% of the training samples as the validation set. That is, during the training procedure, we use 90\% of the training samples to learn network parameters.
\par
Our model is implemented on TensorFlow-1.12.0 and trained for 100 epochs. The computational resource is an NVIDIA Tesla P100 GPU with a 16GB memory. As a compromise between the training speed and GPU memory capacities, we set the size of training batches as 32. To avoid overfitting, the training progress is terminated once the validation loss increases continuously in five epochs.

\subsection{Experimental Setup}
To fully explore the capacity of our proposed network, we extend our researches by replacing the backbone with GoogLeNet (Inceptionv3) \cite{szegedy2015going} and ResNet (ResNet-50 in our case) \cite{he2016deep}. Specifically, we adapt GoogLeNet by removing global average pooling and fully-connected layers as well as reducing the stride of convolutional and pooling layers in “mixed8” to 1 to improve the spatial resolution. Besides, in order to preserve receptive fields of subsequent convolutional layers, filters in “mixed9” are replaced with atrous convolutional filters, and the dilation rate is defined as 2. Regarding ResNet, we set the convolution stride and dilation rate of filters as 1 and 2, respectively, in the last residual block. Global average pooling and fully-connected layers are removed as well.

In our experiments, we compare the proposed attention-aware label relational reasoning network (AL-RN-CNN) with the following competitors: a standard CNN, CNN-RBFNN \cite{zeggada2017deep}, and CA-CNN-BiLSTM \cite{hua2019recurrently}. Regarding the CNN, we replace its last softmax layer, designed for single-label classification, with a sigmoid layer to produce multi-hot sequences. For the CA-CNN-BiLSTM, we follow the experimental configurations in \cite{hua2019recurrently}. Specifically, we first initialize the feature extraction module of CA-CNN-BiLSTM and weights in the bidirectional LSTM layer with CNNs pre-trained on ImageNet dataset and random values from -0.1 to 0.1, respectively. Afterwards, we fine-tune the entire network in the training phase with Nestro Adam optimizer, and the initial learning rate is set as $1e-04$. The loss is calculated with the binary cross-entropy, and the size of training batches is 32. Notably, for all models, output sequences are binarized with a threshold of 0.5 to generate final predictions.

\subsection{Results on the UCM Multi-label Dataset}
\label{sec:out_ucm}

\subsubsection{Quantitative analysis}
\label{sec:ucm_ex1}
In our experiment, we employ $F_1$ \cite{wu2016unified} and $F_2$ \cite{kaggleplanet} scores as evaluation metrics to quantitatively assess the performance of different models. Specifically, these two $F$ scores are calculated with the following equation:
\begin{equation}
\label{eq:f_score}
F_\beta = (1+\beta^2)\frac{p_er_e}{\beta^2p_e+r_e}, \hspace{1em} \beta=1, 2,
\end{equation}
where $p_e$ indicates the example-based precision and recall \cite{tsoumakas2007random} of predictions. Formulas of calculating $p_e$ and $r_e$ are:
\begin{equation}
p_e = \frac{TP_e}{TP_e+FP_e},\hspace{1em} r_e = \frac{TP_e}{TP_e+FN_e},
\end{equation}
where $TP_e$ (example-based true positive) indicates the number of correctly predicted positive labels in an example, while $FP_e$ (example-based false positive) denotes the number of those failed to be recognized. Besides, $FN_e$ (example-based false negative) represents the number of incorrectly predicted negative labels in an example. Here, an example stands for an aerial image and its associated multiple labels. 

To evaluate our network comprehensively, we take mean $F_1$ and $F_2$ score as principal indexes. Moreover, we also report mean $p_e$ and mean $r_e$. In addition to the example-based perspective, label-based precision and recall are also considered and calculated with:
\begin{equation}
\label{eq:m_label}
p_l = \frac{TP_l}{TP_l+FP_l},\hspace{1em} r_l = \frac{TP_l}{TP_l+FN_l},
\end{equation}
to demonstrate the performance of networks from the perspective of each object label.

\begin{table*}[t]
\renewcommand{\arraystretch}{1.3}
\caption{Comparisons of the classification performance on UCM Multi-label Dataset (\%).}
\label{tab:ucm}
\centering
\begin{tabular}{ccccccc}
\Xhline{3\arrayrulewidth}
Network & mean $F_1$ & mean $F_2$ & mean $p_e$ & mean $r_e$ & mean $p_l$ & mean $r_l$\\
\hline
VGGNet \cite{simonyan2014very} & 78.54 & 80.17 & 79.06 & 82.30 & 86.02 & 80.21 \\
VGG-RBFNN \cite{zeggada2017deep} & 78.80 & 81.14 & 78.18 & 83.91 & 81.90 & 82.63\\
CA-VGG-BiLSTM \cite{hua2019recurrently} & 79.78 & 81.69 & 79.33 & 83.99 & 85.28 & 76.52 \\
AL-RN-VGGNet & 85.70 & 85.81 & 87.62 & 86.41 & 91.04 & 81.71 \\
\hline
\hline
GoogLeNet \cite{szegedy2015going} & 80.68 & 82.32 & 80.51 & 84.27 & 87.51 & 80.85 \\
GoogLeNet-RBFNN \cite{zeggada2017deep} & 81.54 & 84.05 & 79.95 & 86.75 & 86.19 & 84.92\\ 
CA-GoogLeNet-BiLSTM \cite{hua2019recurrently} & 81.82 & 84.41 & 79.91 & 87.06 & 86.29 & 84.38\\
AL-RN-GoogLeNet & 85.24 & 85.33 & 87.18 & 85.86 & 91.03 & 81.64 \\
\hline
\hline
ResNet-50 \cite{he2016deep} & 79.68 & 80.58 & 80.86 & 81.95 & 88.78 & 78.98 \\
ResNet-RBFNN \cite{zeggada2017deep} & 80.58 & 82.47 & 79.92 & 84.59 & 86.21 & 83.72 \\
CA-ResNet-BiLSTM \cite{hua2019recurrently} & 81.47 & 85.27 & 77.94 & 89.02 & 86.12 & 84.26 \\
AL-RN-ResNet & 86.76 & 86.67 & 88.81 & 87.07 & 92.33 & 85.95 \\
\Xhline{3\arrayrulewidth}
\end{tabular}

\end{table*}

\begin{table*}
\caption{Example Images and Predicted labels on the UCM and AID Multi-label Dataset.}
\label{tab:predictions}
\centering
\renewcommand{\arraystretch}{1.3}
\begin{threeparttable}
\begin{tabular}{>{\centering\arraybackslash}m{2.5cm}>{\centering\arraybackslash}m{2.6cm}>{\centering\arraybackslash}m{2.6cm}>{\centering\arraybackslash}m{2.6cm}>{\centering\arraybackslash}m{2.6cm}>{\centering\arraybackslash}m{2.6cm}} 
\Xhline{3\arrayrulewidth}
Samples from the UCM Multi-label Dataset 
& \subfloat{\includegraphics[width=0.145\textwidth]{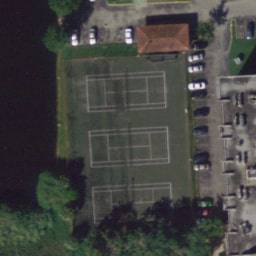}}
& \subfloat{\includegraphics[width=0.145\textwidth]{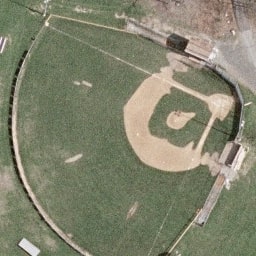}}
& \subfloat{\includegraphics[width=0.145\textwidth]{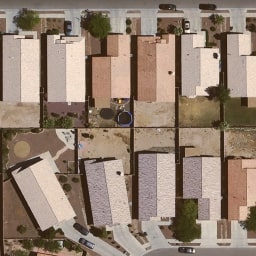}}
& \subfloat{\includegraphics[width=0.145\textwidth]{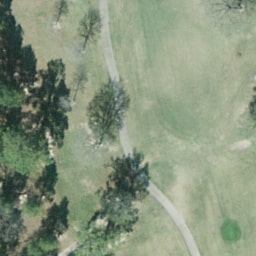}}
& \subfloat{\includegraphics[width=0.145\textwidth]{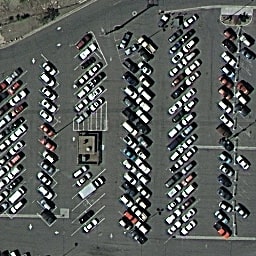}} \\
\hline
Ground Truths & building, car, court, grass, tree, and pavement & building, bare soil, pavement, and grass & car, tree, building, grass, and bare soil  & pavement, grass, tree, and bare soil & car, pavement, and building \\
\hline
Predictions &  building, car, court, grass, tree, and pavement & building, bare soil, pavement, and grass & tree, car, building, grass, bare soil, and \textcolor{red}{pavement} & pavement, grass, tree, and bare soil & car, pavement, and \textcolor{blue}{building} \\
\hline 
\hline
Samples from the AID Multi-label Dataset 
& \subfloat{\includegraphics[width=0.145\textwidth]{example_pred/mediumresidential_72.jpg}}
& \subfloat{\includegraphics[width=0.145\textwidth]{example_pred/playground_87.jpg}}
& \subfloat{\includegraphics[width=0.145\textwidth]{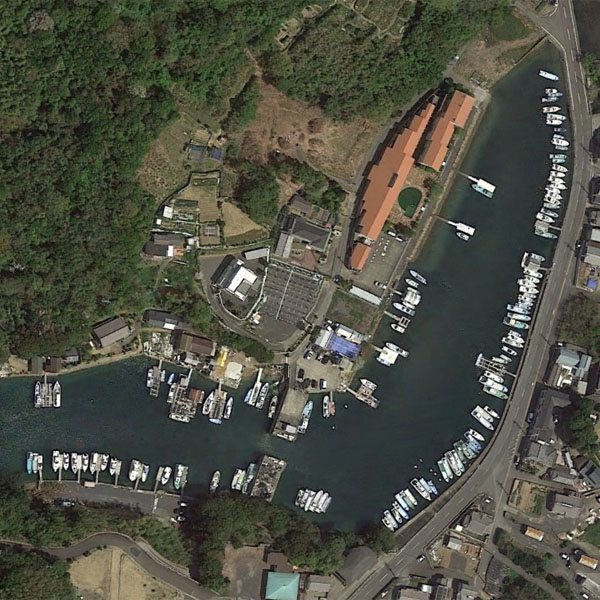}}
& \subfloat{\includegraphics[width=0.145\textwidth]{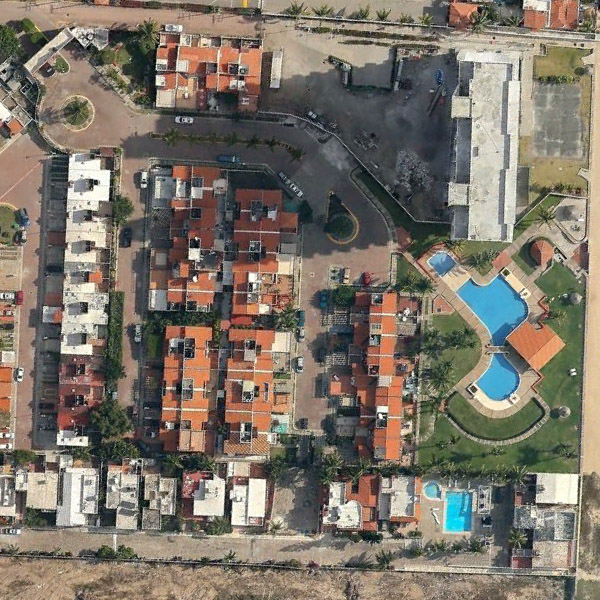}}
& \subfloat{\includegraphics[width=0.145\textwidth]{example_pred/stadium_85.jpg}} \\
\hline
Ground Truths & building, car, grass, tree, and pavement & car, bare soil, court, building, grass, tree, pavement, and water & building, car, tree, dock, grass, pavement, sea, and ship & bare soil, building, car, pavement, grass, tree, and water & court, building, car, bare soil, grass, tree, and pavement \\
\hline
Predictions & building, car, grass, tree, and pavement & car, bare soil, court, building, grass, tree, pavement, and \textcolor{blue}{water} & building, car, tree, dock, grass, pavement, sea, \textcolor{red}{water}, and ship & bare soil, car, building, pavement, water, \textcolor{red}{sand}, tree, and grass & court, building, car, bare soil, grass, tree, and pavement \\
\Xhline{3\arrayrulewidth}
\end{tabular}
\begin{tablenotes}
\item[] \textcolor{red}{Red} predictions indicate false positives, while \textcolor{blue}{blue} predictions are false negatives.
\end{tablenotes}
\end{threeparttable}
\end{table*}

\begin{figure*}[!t]
\captionsetup[subfigure]{captionskip=0.1em, labelformat=empty}
\centering
\subfloat{\includegraphics[width=0.128\textwidth]{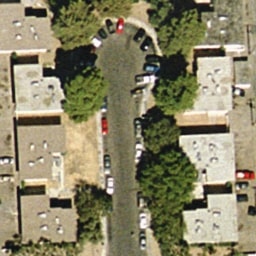}
\label{fig:exs1}}
\hspace{-0.5em}
\subfloat{\includegraphics[width=0.128\textwidth]{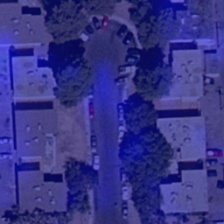}
\label{fig:exs2}}
\hspace{-0.5em}
\subfloat{\includegraphics[width=0.128\textwidth]{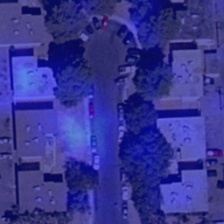}
\label{fig:exs3}}
\hspace{-0.5em}
\subfloat{\includegraphics[width=0.128\textwidth]{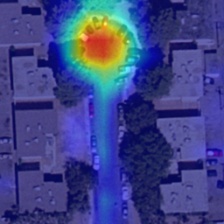}
\label{fig:exs4}}
\hspace{-0.5em}
\subfloat{\includegraphics[width=0.128\textwidth]{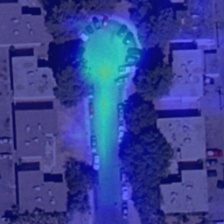}
\label{fig:exs5}}
\hspace{-0.5em}
\subfloat{\includegraphics[width=0.128\textwidth]{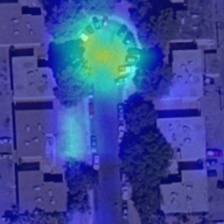}
\label{fig:exs6}}
\hspace{-0.5em}
\subfloat{\includegraphics[width=0.128\textwidth]{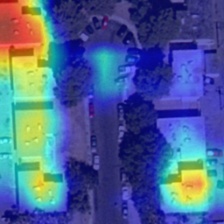}
\label{fig:exs7}}
\vspace{-0.7em}
\subfloat{\includegraphics[width=0.128\textwidth]{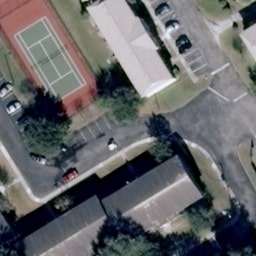}
\label{fig:exs8}}
\hspace{-0.5em}
\subfloat{\includegraphics[width=0.128\textwidth]{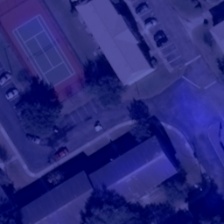}
\label{fig:exs9}}
\hspace{-0.5em}
\subfloat{\includegraphics[width=0.128\textwidth]{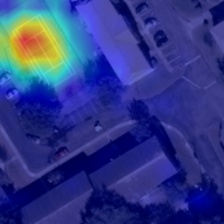}
\label{fig:exs10}}
\hspace{-0.5em}
\subfloat{\includegraphics[width=0.128\textwidth]{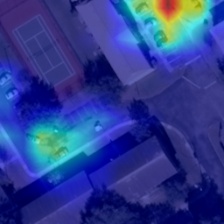}
\label{fig:exs11}}
\hspace{-0.5em}
\subfloat{\includegraphics[width=0.128\textwidth]{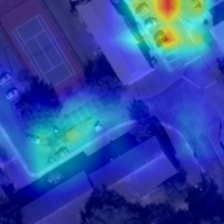}
\label{fig:exs12}}
\hspace{-0.5em}
\subfloat{\includegraphics[width=0.128\textwidth]{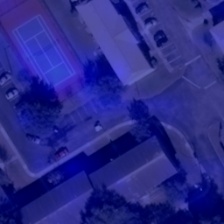}
\label{fig:exs13}}
\hspace{-0.5em}
\subfloat{\includegraphics[width=0.128\textwidth]{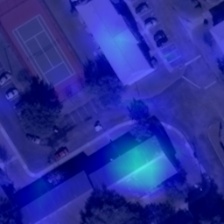}
\label{fig:exs14}}
\vspace{-0.7em}
\subfloat{\includegraphics[width=0.128\textwidth]{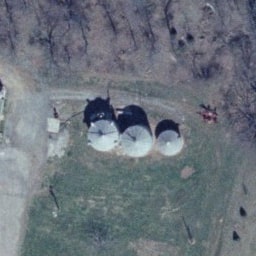}
\label{fig:exs15}}
\hspace{-0.5em}
\subfloat{\includegraphics[width=0.128\textwidth]{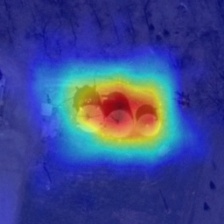}
\label{fig:exs16}}
\hspace{-0.5em}
\subfloat{\includegraphics[width=0.128\textwidth]{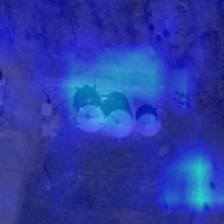}
\label{fig:exs17}}
\hspace{-0.5em}
\subfloat{\includegraphics[width=0.128\textwidth]{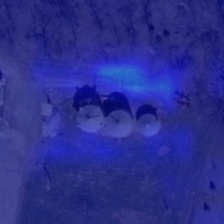}
\label{fig:exs18}}
\hspace{-0.5em}
\subfloat{\includegraphics[width=0.128\textwidth]{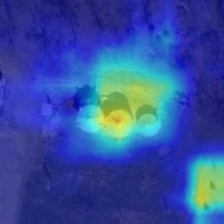}
\label{fig:exs19}}
\hspace{-0.5em}
\subfloat{\includegraphics[width=0.128\textwidth]{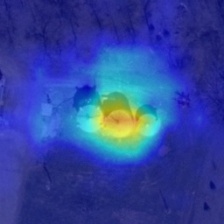}
\label{fig:exs20}}
\hspace{-0.5em}
\subfloat{\includegraphics[width=0.128\textwidth]{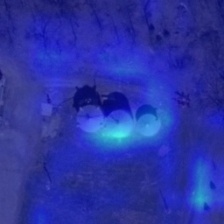}
\label{fig:exs21}}
\vspace{-0.7em}
\subfloat[(a)]{\includegraphics[width=0.128\textwidth]{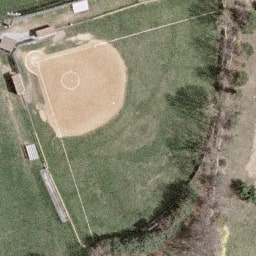}
\label{fig:exs22}}
\hspace{-0.5em}
\subfloat[(b)]{\includegraphics[width=0.128\textwidth]{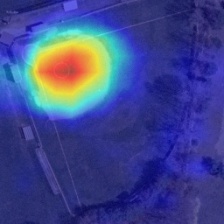}
\label{fig:exs23}}
\hspace{-0.5em}
\subfloat[(c)]{\includegraphics[width=0.128\textwidth]{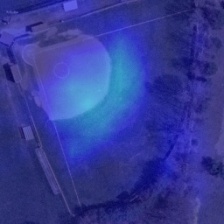}
\label{fig:exs24}}
\hspace{-0.5em}
\subfloat[(d)]{\includegraphics[width=0.128\textwidth]{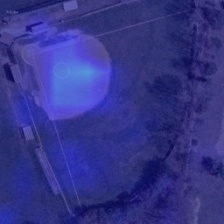}
\label{fig:exs25}}
\hspace{-0.5em}
\subfloat[(e)]{\includegraphics[width=0.128\textwidth]{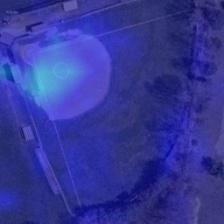}
\label{fig:exs26}}
\hspace{-0.5em}
\subfloat[(f)]{\includegraphics[width=0.128\textwidth]{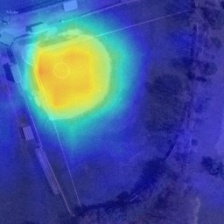}
\label{fig:exs27}}
\hspace{-0.5em}
\subfloat[(g)]{\includegraphics[width=0.128\textwidth]{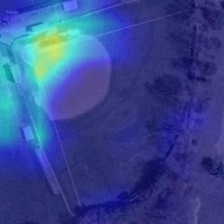}
\label{fig:exs28}}
\caption{Example label-specific features of (a) samples selected from the UCM multi-label dataset regarding (b) tank, (c) court, (d) pavement, (e) car, (f) bare soil, and (g) building. Red implies strong activations, while blue indicates weak activations.}
\label{fig:featureU}
\end{figure*}
\begin{figure*}[!t]
\centering
\subfloat[]{\includegraphics[width=0.22\textwidth]{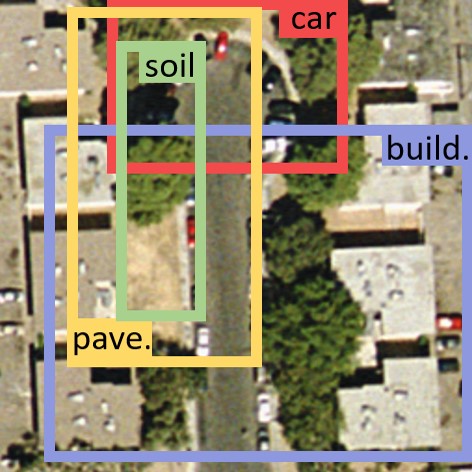}
\label{fig:roi_ua}}
\hfil
\subfloat[]{\includegraphics[width=0.22\textwidth]{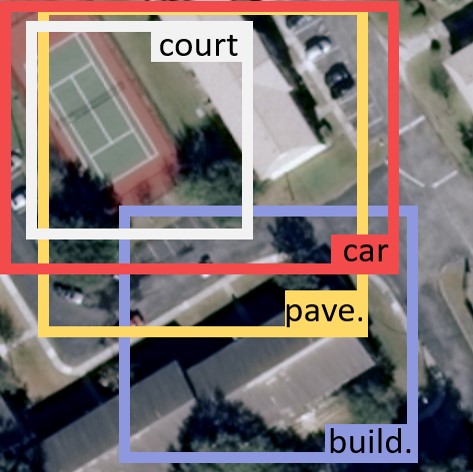}
\label{fig:roi_ub}}
\hfil
\subfloat[]{\includegraphics[width=0.22\textwidth]{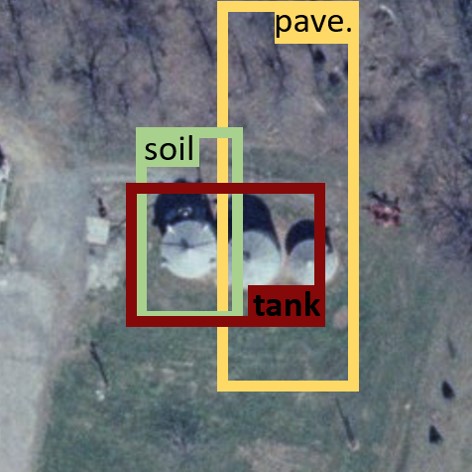}
\label{fig:roi_uc}}
\hfil
\subfloat[]{\includegraphics[width=0.22\textwidth]{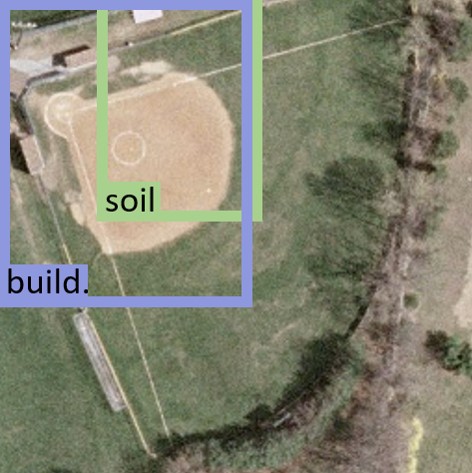}
\label{fig:roi_ud}}
\caption{Example attentional regions for car, bare soil (soil), building (build.), pavement (pave.), court, and tank in various scenes (a)-(d) in the UCM multi-label dataset. For each scene, only positive labels mentioned in Fig.~\ref{fig:featureU} are considered.}
\label{fig:roiU}
\end{figure*}

\begin{figure*}[!t]
\centering
\subfloat[]{\includegraphics[width=0.232\textwidth]{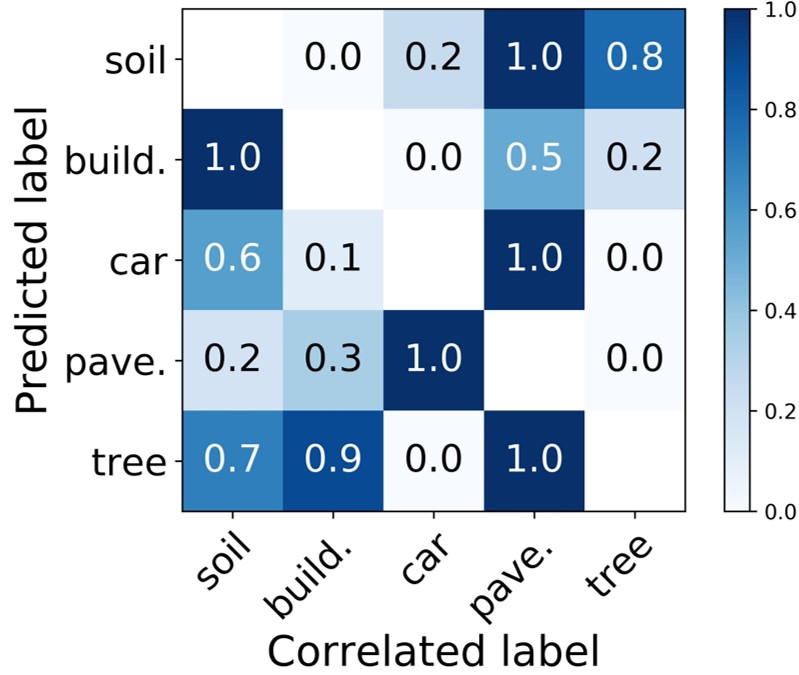}
\label{fig:relation_ua}}
\hfil
\subfloat[]{\includegraphics[width=0.232\textwidth]{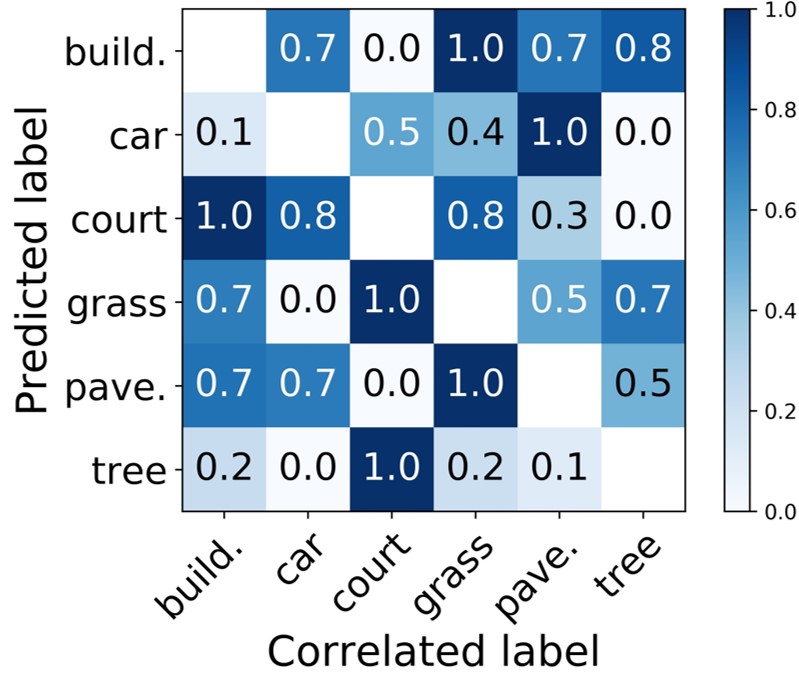}
\label{fig:relation_ub}}
\hfil
\subfloat[]{\includegraphics[width=0.232\textwidth]{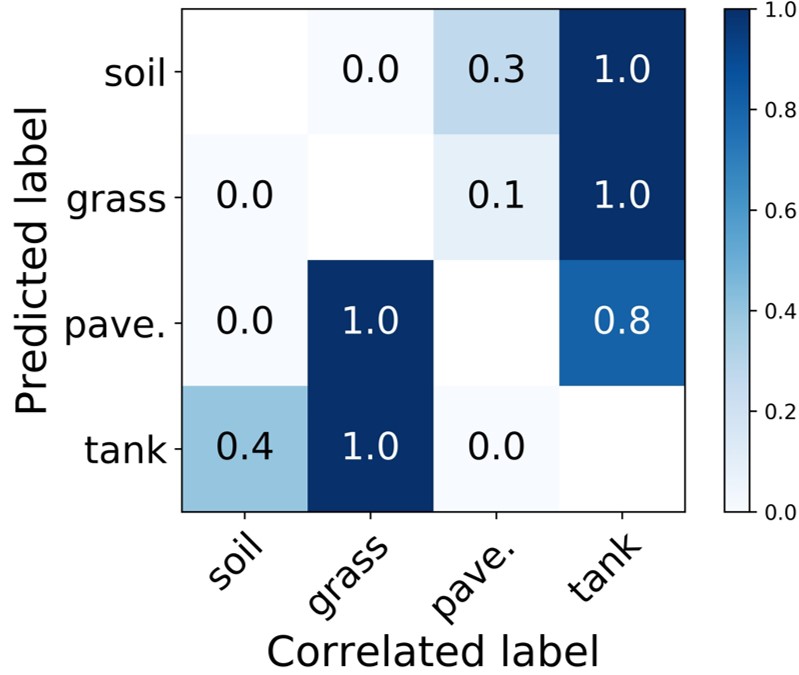}
\label{fig:relation_uc}}
\hfil
\subfloat[]{\includegraphics[width=0.232\textwidth]{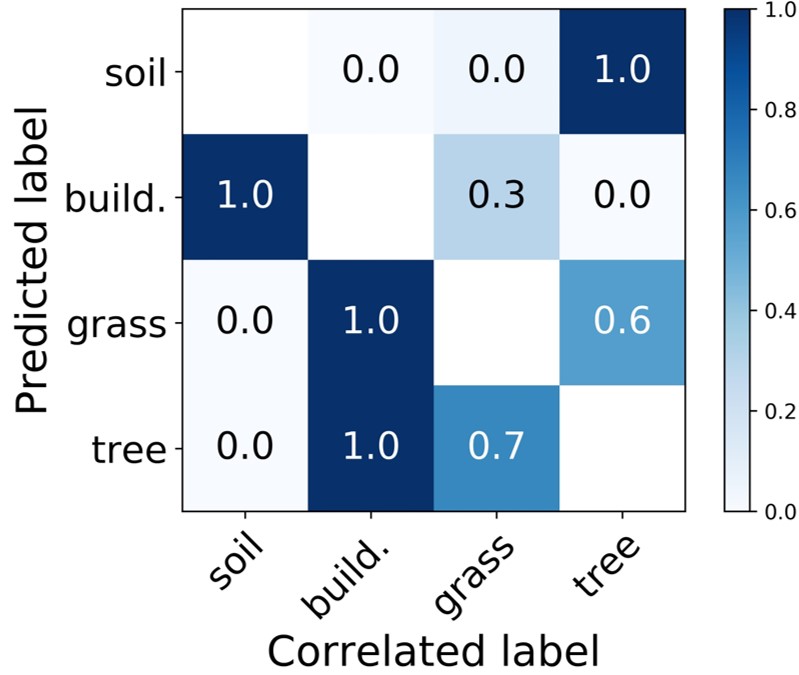}
\label{fig:relation_ud}}
\caption{Example pairwise relations among labels present in scene (a)-(d), which are shown in Fig.~\ref{fig:roiU}. Each label at Y-axis represents the predicted label $l$, and labels at X-axis are correlated labels. Normalization is performed according to each row, and white color represents null values.}
\label{fig:relation_matrixU}
\end{figure*}

\begin{table*}[t]
\renewcommand{\arraystretch}{1.3}
\caption{Comparisons of the classification performance on AID Multi-label Dataset (\%).}
\label{tab:aid}
\centering
\begin{tabular}{ccccccc}
\Xhline{3\arrayrulewidth}
Network & mean $F_1$ & mean $F_2$ & mean $p_e$ & mean $r_e$ & mean $p_l$ & mean $r_l$\\
\hline
VGGNet \cite{simonyan2014very} & 85.52 & 85.60 & 87.41 & 86.32 & 70.60 & 58.89 \\
VGG-RBFNN \cite{zeggada2017deep} & 84.58 & 85.99 & 84.56 & 87.85 & 62.90 & 69.15\\
CA-VGG-BiLSTM \cite{hua2019recurrently} & 86.68 & 86.88 & 88.68 & 87.83 & 72.04 & 60.00 \\
proposed AL-RN-VGGNet & 88.09 & 88.31 & 89.96 & 89.27 & 76.94 & 68.31 \\
\hline
\hline
GoogLeNet \cite{szegedy2015going} & 86.27 & 85.77 & 89.49 & 86.00 & 74.18 & 53.69 \\
GoogLeNet-RBFNN \cite{zeggada2017deep} & 84.85 & 86.80 & 84.68 & 89.14 & 65.41 & 72.26 \\ 
CA-GoogLeNet-BiLSTM \cite{hua2019recurrently} & 85.36 & 85.21 & 88.05 & 85.79 & 68.80 & 59.36 \\
proposed AL-RN-GoogLeNet & 88.17 & 88.25 & 90.03 & 88.77 & 77.92 & 69.50 \\
\hline
\hline
ResNet-50 \cite{he2016deep} & 86.23 & 85.57 & 89.31 & 85.65 & 72.39 & 52.82 \\
ResNet-RBFNN \cite{zeggada2017deep} & 83.77 & 85.87 & 82.84 & 88.32 & 60.85 & 70.45 \\
CA-ResNet-BiLSTM \cite{hua2019recurrently} & 87.63 & 88.03 & 89.03 & 88.99 & 79.50 & 65.60 \\
proposed AL-RN-ResNet & 88.72 & 88.54 & 91.00 & 88.95 & 80.81 & 71.12 \\
\Xhline{3\arrayrulewidth}
\end{tabular}
\end{table*}

\begin{figure*}[!t]
\captionsetup[subfigure]{captionskip=0.1em, labelformat=empty}
\centering
\subfloat{\includegraphics[width=0.128\textwidth]{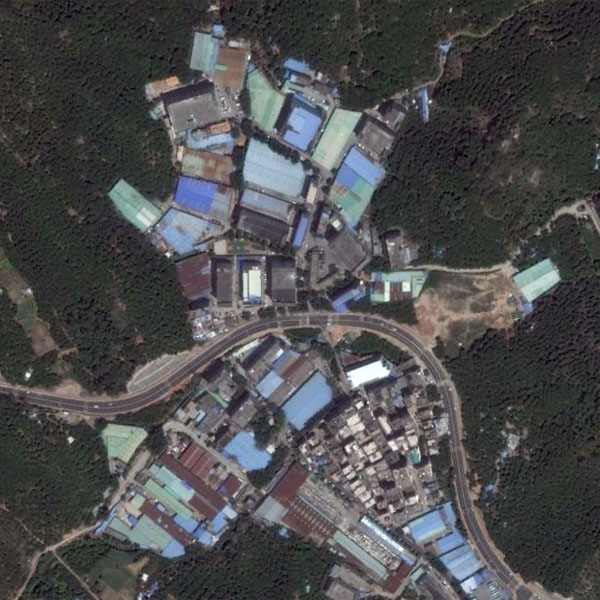}
\label{fig:aidexs1}}
\hspace{-0.5em}
\subfloat{\includegraphics[width=0.128\textwidth]{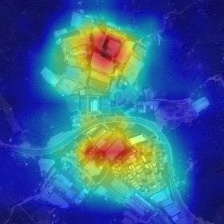}
\label{fig:aidexs2}}
\hspace{-0.5em}
\subfloat{\includegraphics[width=0.128\textwidth]{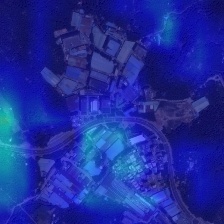}
\label{fig:aidexs3}}
\hspace{-0.5em}
\subfloat{\includegraphics[width=0.128\textwidth]{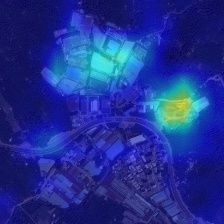}
\label{fig:aidexs4}}
\hspace{-0.5em}
\subfloat{\includegraphics[width=0.128\textwidth]{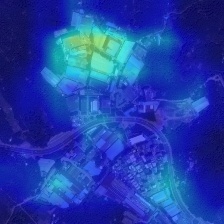}
\label{fig:aidexs5}}
\hspace{-0.5em}
\subfloat{\includegraphics[width=0.128\textwidth]{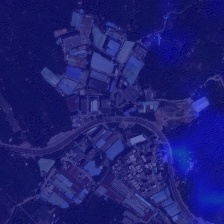}
\label{fig:aidexs6}}
\hspace{-0.5em}
\subfloat{\includegraphics[width=0.128\textwidth]{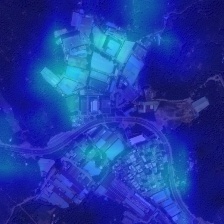}
\label{fig:aidexs7}}
\vspace{-0.7em}
\subfloat{\includegraphics[width=0.128\textwidth]{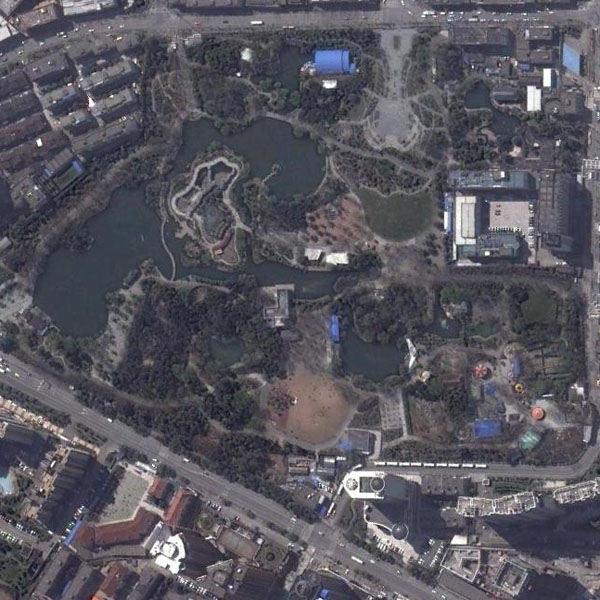}
\label{fig:aidexs8}}
\hspace{-0.5em}
\subfloat{\includegraphics[width=0.128\textwidth]{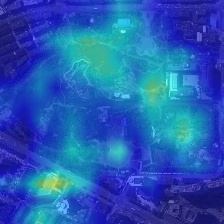}
\label{fig:aidexs9}}
\hspace{-0.5em}
\subfloat{\includegraphics[width=0.128\textwidth]{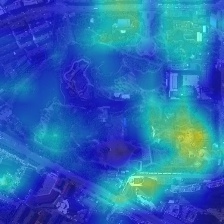}
\label{fig:aidexs10}}
\hspace{-0.5em}
\subfloat{\includegraphics[width=0.128\textwidth]{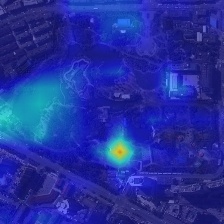}
\label{fig:aidexs11}}
\hspace{-0.5em}
\subfloat{\includegraphics[width=0.128\textwidth]{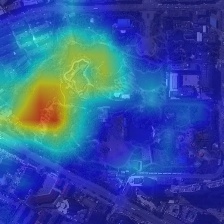}
\label{fig:aidexs12}}
\hspace{-0.5em}
\subfloat{\includegraphics[width=0.128\textwidth]{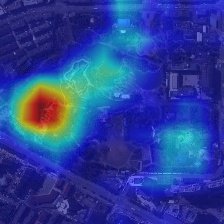}
\label{fig:aidexs13}}
\hspace{-0.5em}
\subfloat{\includegraphics[width=0.128\textwidth]{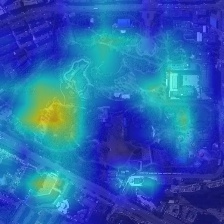}
\label{fig:aidexs14}}
\vspace{-0.7em}
\subfloat{\includegraphics[width=0.128\textwidth]{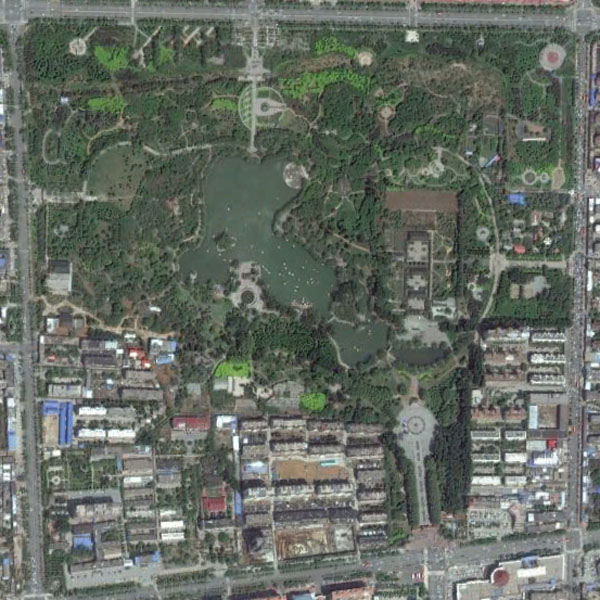}
\label{fig:aidexs15}}
\hspace{-0.5em}
\subfloat{\includegraphics[width=0.128\textwidth]{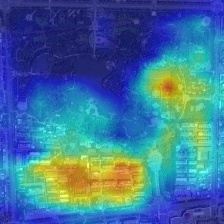}
\label{fig:aidexs16}}
\hspace{-0.5em}
\subfloat{\includegraphics[width=0.128\textwidth]{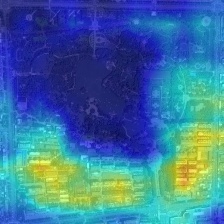}
\label{fig:aidexs17}}
\hspace{-0.5em}
\subfloat{\includegraphics[width=0.128\textwidth]{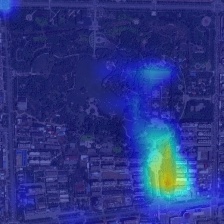}
\label{fig:aidexs18}}
\hspace{-0.5em}
\subfloat{\includegraphics[width=0.128\textwidth]{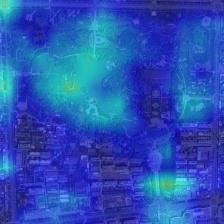}
\label{fig:aidexs19}}
\hspace{-0.5em}
\subfloat{\includegraphics[width=0.128\textwidth]{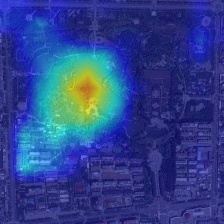}
\label{fig:aidexs20}}
\hspace{-0.5em}
\subfloat{\includegraphics[width=0.128\textwidth]{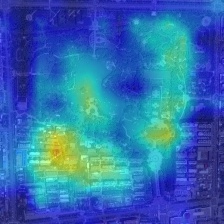}
\label{fig:aidexs21}}
\vspace{-0.7em}
\subfloat[(a)]{\includegraphics[width=0.128\textwidth]{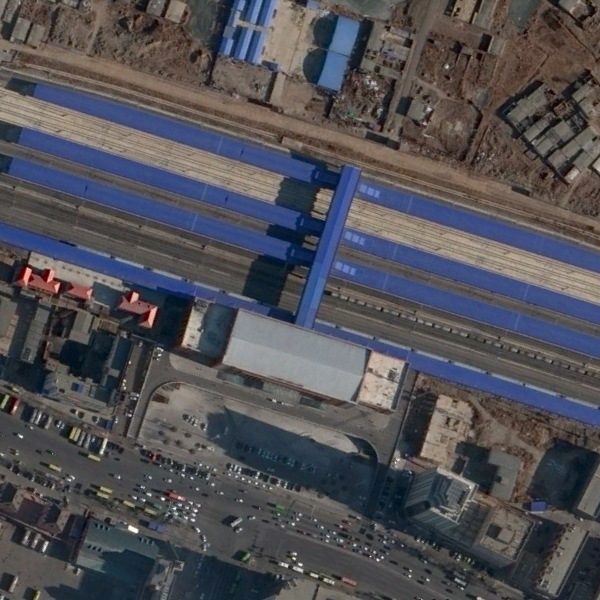}
\label{fig:aidexs22}}
\hspace{-0.5em}
\subfloat[(b)]{\includegraphics[width=0.128\textwidth]{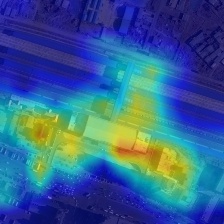}
\label{fig:aidexs23}}
\hspace{-0.5em}
\subfloat[(c)]{\includegraphics[width=0.128\textwidth]{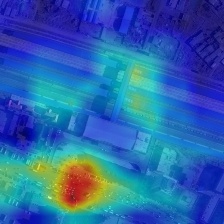}
\label{fig:aidexs24}}
\hspace{-0.5em}
\subfloat[(d)]{\includegraphics[width=0.128\textwidth]{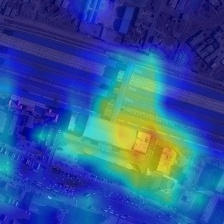}
\label{fig:aidexs25}}
\hspace{-0.5em}
\subfloat[(e)]{\includegraphics[width=0.128\textwidth]{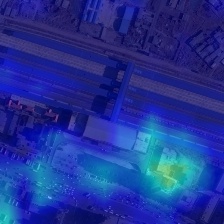}
\label{fig:aidexs26}}
\hspace{-0.5em}
\subfloat[(f)]{\includegraphics[width=0.128\textwidth]{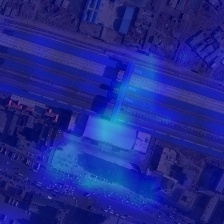}
\label{fig:aidexs27}}
\hspace{-0.5em}
\subfloat[(g)]{\includegraphics[width=0.128\textwidth]{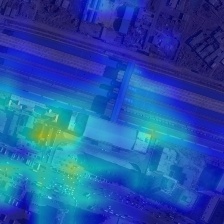}
\label{fig:aidexs28}}
\caption{Example label-specific features of (a) samples selected from the AID multi-label dataset regarding (b) building, (c) car, (d) bare soil, (e) tree, (f) water, and (g) pavement. Red implies strong activations, while blue indicates weak activations.}
\label{fig:featureA}
\end{figure*}

Table \ref{tab:ucm} exhibits experimental results on the UCM multi-label dataset. We can observe that our model surpasses all competitors on the UCM multi-label dataset with variant backbones. Specifically, AL-RN-VGGNet increases mean $F_1$ and $F_2$ scores by 7.16\% and 5.64\%, respectively, in comparison with VGGNet. Compared to CA-VGG-BiLSTM, which resorts to employing a bidirectional LSTM structure for exploring label dependencies, our network obtains an improvement of 5.92\% in the mean $F_1$ score. Besides, although CA-VGG-BiLSTM is superior to VGGNet in both mean $F_1$ and $F_2$ scores, it achieves decreased mean precisions and recalls. In contrast, AL-RN-VGGNet outperforms VGGNet not only in mean $F_1$ and $F_2$ scores but also in mean example- and label-based precisions and recalls. For another backbone, GoogLeNet, our network gains the best mean $F_1$ and $F_2$ scores. As shown in Table \ref{tab:ucm}, AL-RN-GoogLeNet increases the mean $F_1$ score by 4.56\% and 3.42\% with respect to GoogLeNet and CA-GoogLeNet-BiLSTM, respectively. For the mean $F_2$ score and precisions, our model also surpasses other competitors, which proves the effectiveness and robustness of our method. AL-RN-ResNet achieves the best mean $F_1$ score, 0.8676, and $F_2$ score, 0.8667, in comparison with all other models. Furthermore, it obtains the best mean example-based precision, 0.8881, and label-based precision, 0.9233, and recall, 0.8595. To summarize, comparisons between AL-RN-CNN and other models demonstrate the effectiveness of our network. Moreover, comparisons between AL-RN-CNN and CA-CNN-BiLSTM illustrate that the composite function-based proposed model performs better than a BiLSTM framework in terms of both accuracy and robustness. Reasons could be that: 1) a chain-like BiLSTM architecture might suffer from the error propagation \cite{Demir2019label} and thus is sensitive to the order of predictions, while in our network, all pair-wise label relations are encoded separately and the final summation function is order invariant \cite{santoro2017simple}. 2) a BiLSTM-based structure models label relations implicitly, whereas our network encodes such relations in an explicit and direct way. Table~\ref{tab:predictions} presents several example predictions from the UCM multi-label dataset. As a supplementary study, we evaluate the robustness of our proposed model by performing cross-validation in the training phase. More specifically, we randomly divide training samples into five folds and train our best-performed model, i.e., AL-RN-ResNet, five times. For each training progress, we select one of five folds as the validation set and train our model with the remaining four folds. We observe that variances of mean $F_1$ and $F_2$ scores are 0.38\% and 0.71\%, respectively. Compared to improvements brought by our network, variances are limited, and this demonstrates the robustness of our proposed network.

\begin{figure*}[!t]
\centering
\subfloat[]{\includegraphics[width=0.22\textwidth]{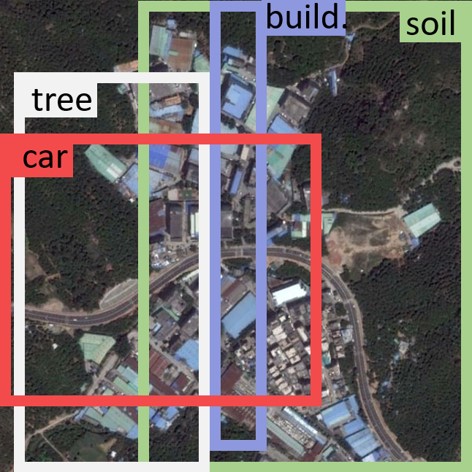}
\label{fig:roi_aa}}
\hfil
\subfloat[]{\includegraphics[width=0.22\textwidth]{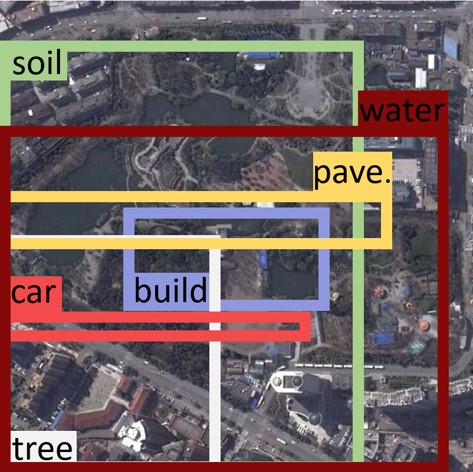}
\label{fig:roi_ab}}
\hfil
\subfloat[]{\includegraphics[width=0.22\textwidth]{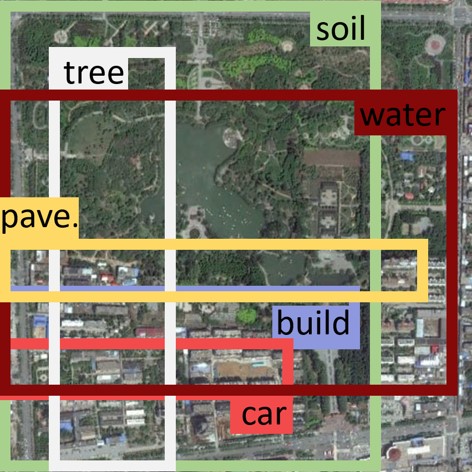}
\label{fig:roi_ac}}
\hfil
\subfloat[]{\includegraphics[width=0.22\textwidth]{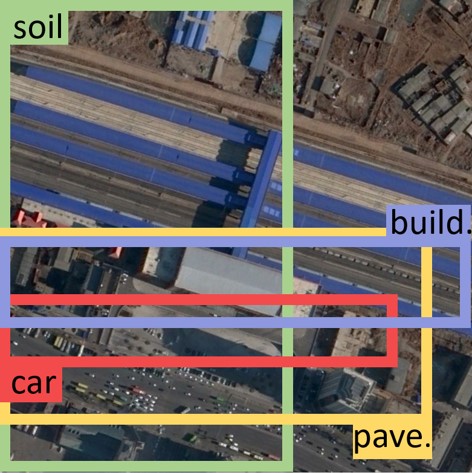}
\label{fig:roi_ad}}
\caption{Example attentional regions for car, bare soil (soil), building (build.), pavement (pave.), court, and tank in various scenes (a)-(d) in the AID multi-label dataset. For each scene, only positive labels mentioned in Fig.~\ref{fig:featureA} are considered.}
\label{fig:roiA}
\end{figure*}

\begin{figure*}[!t]
\centering
\subfloat[]{\includegraphics[width=0.232\textwidth]{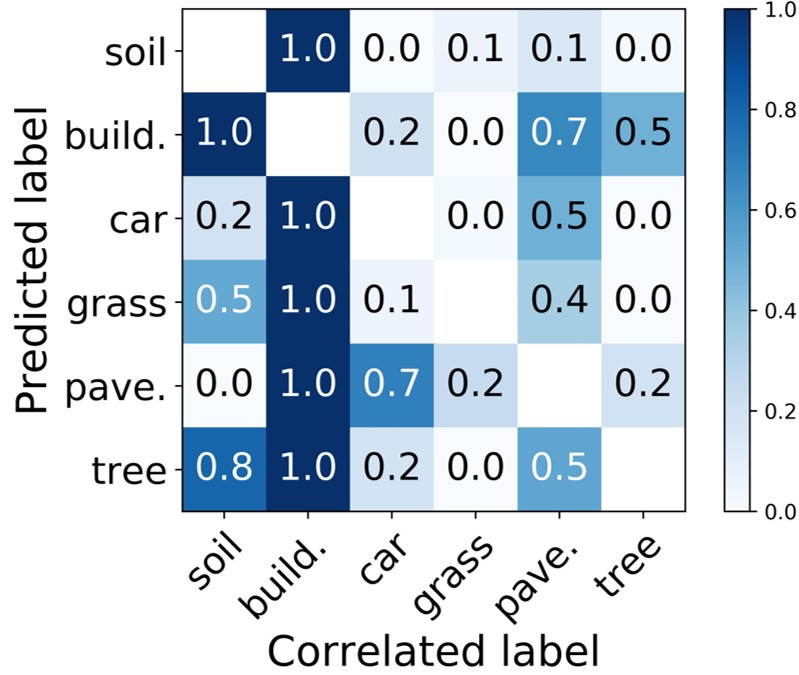}
\label{fig:relation_aa}}
\hfil
\subfloat[]{\includegraphics[width=0.232\textwidth]{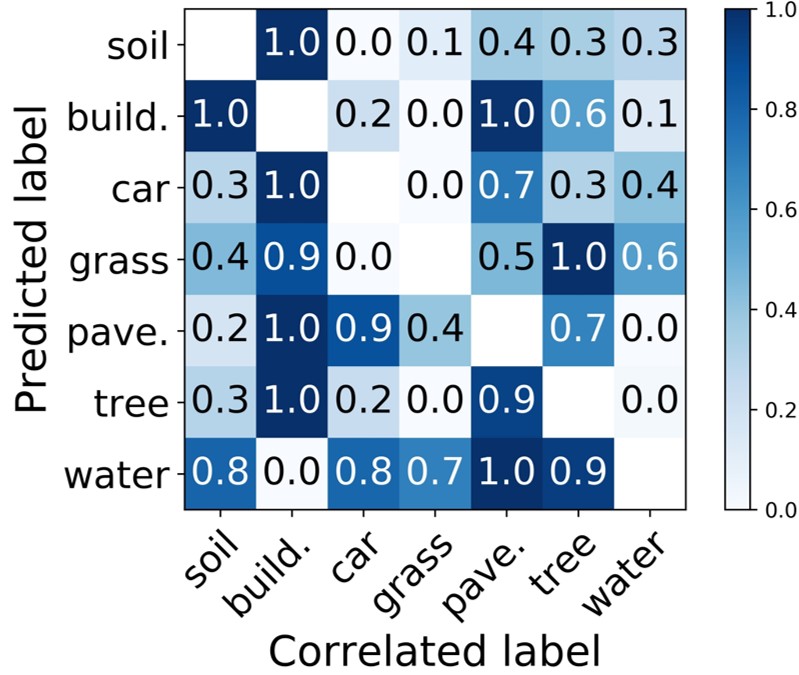}
\label{fig:relation_ab}}
\hfil
\subfloat[]{\includegraphics[width=0.232\textwidth]{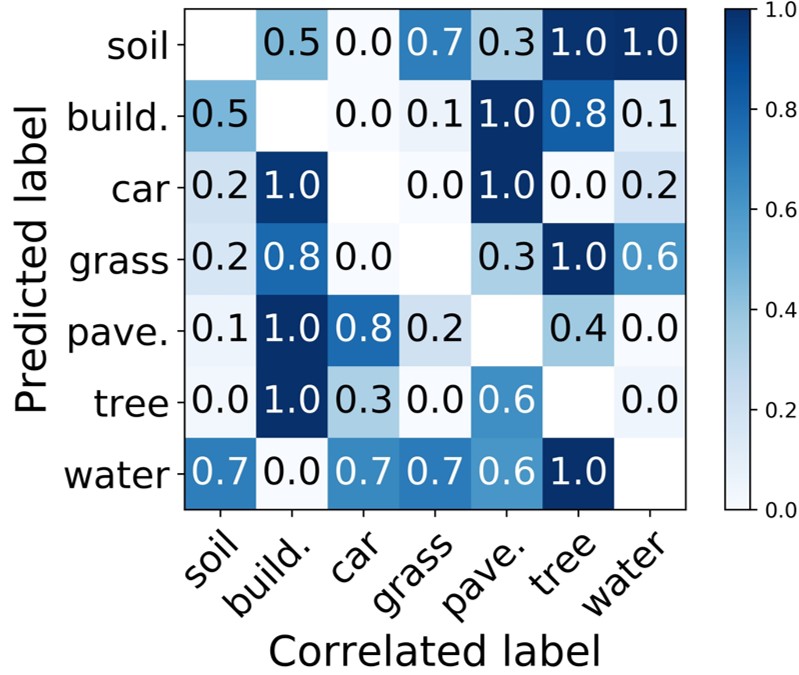}
\label{fig:relation_ac}}
\hfil
\subfloat[]{\includegraphics[width=0.232\textwidth]{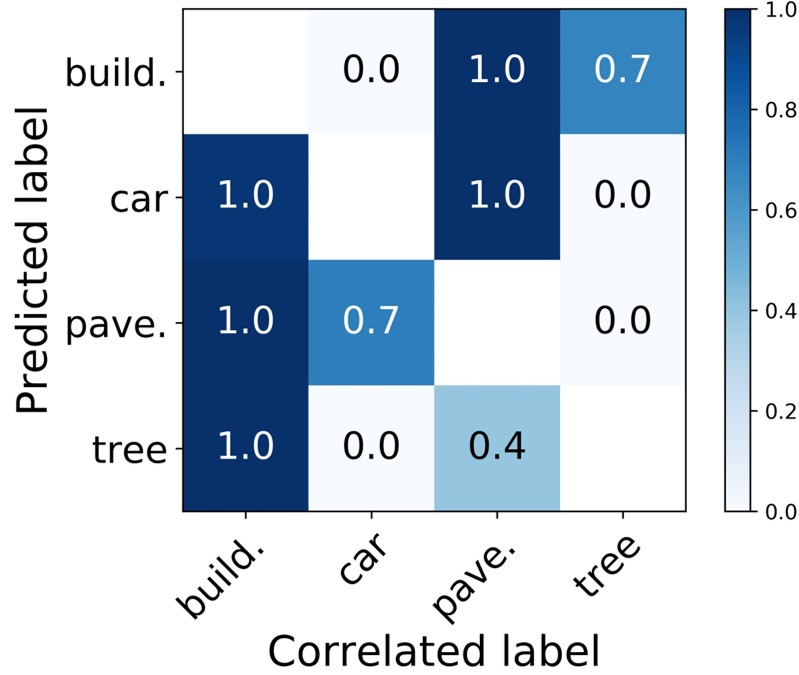}
\label{fig:relation_ad}}
\caption{Example pairwise relations among labels present in scene (a)-(d), which are shown in Fig.~\ref{fig:roiA}. Each label at Y-axis represents the predicted label $l$, and labels at X-axis are correlated labels. Normalization is performed according to each row, and white color represents null values.}
\label{fig:relation_matrixA}
\end{figure*}

\subsubsection{Qualitative analysis}
In order to figure out what is going on inside our network, we further visualize features learned from each module and validate the effectiveness of the proposed network in a qualitative manner. In Fig.~\ref{fig:featureU}, a couple of feature parcels regarding bare soil, building, car, pavement, court, and tank is displayed for several example images. Note that for $K$ feature maps in each feature parcel, we select the most strongly activated one as the representative. We can observe that discriminative regions related to positive labels are highlighted in these feature maps, while less informative regions are weakly activated. As an exception, the feature map at the bottom left of Fig.~\ref{fig:featureU} shows that the baseball field is misidentified as tanks, which may lead to incorrect predictions.

\par
For evaluating the localization ability of the proposed network, we visualize attentional regions learned from the second module. Coordinates of bottom left (BL) and top right (TR) corners of attentional region grids are calculated with the following equation:
\begin{equation}
\label{eq:st_corner}
    \begin{bmatrix}
        x^{attn}_{BL} & x^{attn}_{TR}\\
        y^{attn}_{BL} & y^{attn}_{TR}
    \end{bmatrix} 
    = 
    \bm{M}_{T_l}
    \begin{bmatrix}
        -1 & 1\\
        -1 & 1\\
        1 & 1
    \end{bmatrix}.    
\end{equation}

Fig.~\ref{fig:roiU} shows some examples of learned attentional regions. As we can see, most attentional regions concentrate on areas covering objects of interest. Besides, it is noteworthy that even objects are distributed dispersedly, the learned attentional regions can still cover most of them, e.g., buildings in Fig.~\ref{fig:roi_ua} and cars in \ref{fig:roi_ub}. 

Furthermore, learned pairwise label relations are visualized in the format of matrix, where an element at $(l, m)$ indicates ${\rm LR}(\bm{A}_{l}, \bm{A}_{m})$. Fig.~\ref{fig:relation_matrixU} exhibits some examples for the four scenes in Fig.~\ref{fig:roiU}. In these examples, we take only positive object labels into consideration and perform normalization alongside each row to yield a distinct visualization of \textit{``label relations''}. Since $m$ differs from $l$, we assign null values to diagonal elements and mark them as white color in Fig.~\ref{fig:relation_matrixU}. It can be seen that in Fig.~\ref{fig:relation_ua} and \ref{fig:relation_ub}, relations between car and pavement contribute significantly to predicting presences of both car and pavement. Besides, Fig.~\ref{fig:relation_ud} shows that the existence of tree highly suggests the presence of bare soil, but not vice versa. These observations illustrate that even without prior knowledge, the proposed network can reason about relations, that are in line with the reality.

\subsection{Results on the AID Multi-label Dataset}
\label{sec:out_aid}
\subsubsection{Quantitative analysis}
To further evaluate the proposed network, we report experimental results on the AID multi-label dataset. Evaluation metrics here are the same as those in previous experiments, and results are presented in Table~\ref{tab:aid}. As we can observe, the proposed AL-RN-CNN behaves superior to all competitors in most of the metrics. To be more specific, AL-RN-VGGNet improves the mean $F_1$ and $F_2$ score by 2.57\% and 2.71\%, respectively, compared to the baseline model. In comparison with CA-VGG-BiLSTM, our network gains an improvement of 1.41\% in the mean $F_1$ score and 1.43\% in the mean $F_2$ score. Regarding the other two backbones, similar phenomena can be observed as well. AL-RN-GoogLeNet achieves the highest mean $F_1$ and $F_2$ score, 0.8817 and 0.8825, compared to GoogLeNet and CA-GoogLeNet-BiLSTM, while AL-RN-ResNet surpasses the second-best model by 1.09\% and 0.51\% in the mean $F_1$ and $F_2$ score, respectively. Besides, it is noteworthy that although CA-GoogLeNet-BiLSTM shows a decreased performance compared to the baseline model, our network still achieves higher scores in all metrics. Moreover, we notice that the proposed AL-RN-CNNs outperform baseline CNNs by a large margin in the mean label-based recall, and the maximum improvement can reach 18.30\%. In conclusion, these comparisons suggest that explicitly modeling label relations can improve the robustness and retrieval ability of a network. Several example predictions on the AID multi-label dataset are presented in Table~\ref{tab:predictions}.

\subsubsection{Qualitative analysis}

To dive deep into the model, we visualize label-specific features and attentional regions in Fig.~\ref{fig:featureA} and \ref{fig:roiA}, respectively. In Fig.~\ref{fig:featureA}, representative feature maps in various feature parcels for bare soil, building, car, pavement, tree, and water are displayed. As shown here, regions with label-related semantics are highlighted, while less informative regions present weak activations. For instance, regions of ponds are considered as discriminative regions for identifying \textsl{water}. Residential and industrial areas are strongly activated in feature maps for recognizing \textsl{building}. In Fig.~\ref{fig:roiA}, it can be observed that attentional regions learned from our network are able to capture areas of semantic objects, such as cars and trees. We also note that some attentional regions in Fig.~\ref{fig:roiA} are coarser than those in Fig.~\ref{fig:roiU}, which is because the AID multi-label dataset has a lower spatial resolution.

Furthermore, pairwise relations among positive labels are visualized in Fig.~\ref{fig:relation_matrixA}. As shown in Fig.~\ref{fig:relation_ab}, \ref{fig:relation_ac}, and \ref{fig:relation_ad}, existences of both tree and pavement contribute significantly to the identification of car, while the occurrence of car only suggests a high probability that pavement presents. Strong pairwise relations between building and other labels, e.g., car, pavement, and tree, indicate that the presence of building can heavily assist in predicting those labels.

\begin{table}[t]
\renewcommand{\arraystretch}{1.3}
\caption{Comparison between different $g_{\theta_{lm}}$ (\%).}
\label{tab:1d2d}
\centering
\begin{threeparttable}
\begin{tabular}{P{1.1cm}|c|P{0.55cm}P{0.55cm}P{0.55cm}|P{0.55cm}P{0.55cm}P{0.55cm}}
\Xhline{3\arrayrulewidth}
Dataset & $g_{\theta_{lm}}$ & $\textnormal{V*}_{F_1}$ & $\textnormal{G*}_{F_1}$ & $\textnormal{R*}_{F_1}$ & $\textnormal{V*}_{F_2}$ & $\textnormal{G*}_{F_2}$ & $\textnormal{R*}_{F_2}$\\
\hline
\multirow{2}{*}{UCM mul.} & MLP & 82.11 & 83.02 & 85.36 & 81.99 & 84.02 & 86.09 \\
& Conv. & 85.70 & 85.24 & 86.76 & 85.81 & 85.33 & 86.67 \\
\hline
\hline
\multirow{2}{*}{AID mul.} & MLP & 87.79 & 84.92 & 87.10 & 87.74 & 86.97 & 86.83 \\
& Conv. & 88.09 & 88.17 & 88.72 & 88.31 & 88.25 & 88.54 \\
\Xhline{3\arrayrulewidth}
\end{tabular}
\begin{tablenotes}
\item[] ${V*}_{F_1}$, ${G*}_{F_1}$, and ${R*}_{F_1}$ indicate the mean $F_1$ score achieved by VGGNet-, GoogLeNet-, and ResNet-based networks.
\item[] ${V*}_{F_2}$, ${G*}_{F_2}$, and ${R*}_{F_2}$ indicate the mean $F_2$ score achieved by VGGNet-, GoogLeNet-, and ResNet-based networks.
\end{tablenotes}
\end{threeparttable}
\end{table}

\subsection{Discussion on the Relational Inference Module}
Regarding the relational inference module, the function $g_{\theta_{lm}}$ is an important component, which reasons about relations between two objects. Hence, in this subsection, we discuss about different implementations of $g_{\theta_{lm}}$. Specifically, we compare our AL-RN-CNN with LR-CNN~\cite{hua2019label}, which employs a global average pooling layer and an MLP as $g_{\theta_{lm}}$, on both the UCM and AID multi-label datasets. Experimental results are reported in Table~\ref{tab:1d2d}. As shown in this table, our network gains the best mean $F_1$ and $F_2$ score on both datasets with variant backbones. AL-RN-VGGNet achieves the highest improvements of 3.59\% and 3.82\% for the mean $F_1$ and $F_2$ score, respectively, compared to LR-VGGNet on the UCM multi-label dataset. AL-RN-GoogLeNet increases the mean $F_1$ and $F_2$ score by 3.25\% and 1.28\%, respectively, in comparison with LR-ResNet on the AID multi-label dataset. Moreover, AL-RN-CNN can encode label relations through various fields of view by simply changing the size of convolutional filters in $g_{\theta_{lm}}$.

\section{Conclusion}
\label{sec:conclusion}

In this work, we propose a novel aerial image multi-label classification network, namely attention-aware label relational reasoning network. This network comprises three components: a label-wise feature parcel learning module, an attentional region extraction module, and a label relational inference module. To be more specific, the label-wise feature parcel learning module is designed to learn high-level feature parcels, which are proven to encompass label-relevant semantics, and the attentional region extraction module further generates finer attentional feature parcels by preserving only features located in discriminative regions. Afterwards, the label relational inference module reasons about pairwise relations among all labels and exploit these relations for the final prediction. In order to assess the performance of our network, experiments are conducted on the UCM multi-label dataset and a newly proposed AID multi-label dataset. In comparison with other deep learning methods, our network can offer better classification results. In addition, we visualize extracted feature parcels, attentional regions, and relation matrices for demonstrating the effectiveness of each module in a qualitative way. Looking into the future, such network architecture has several potentials, e.g., weakly supervised object detection and semantic segmentation.


\ifCLASSOPTIONcaptionsoff
  \newpage
\fi

\bibliographystyle{IEEEtran}
\bibliography{reference.bib}

\begin{thebibliography}{10}
\providecommand{\url}[1]{#1}
\csname url@samestyle\endcsname
\providecommand{\newblock}{\relax}
\providecommand{\bibinfo}[2]{#2}
\providecommand{\BIBentrySTDinterwordspacing}{\spaceskip=0pt\relax}
\providecommand{\BIBentryALTinterwordstretchfactor}{4}
\providecommand{\BIBentryALTinterwordspacing}{\spaceskip=\fontdimen2\font plus
\BIBentryALTinterwordstretchfactor\fontdimen3\font minus
  \fontdimen4\font\relax}
\providecommand{\BIBforeignlanguage}[2]{{%
\expandafter\ifx\csname l@#1\endcsname\relax
\typeout{** WARNING: IEEEtran.bst: No hyphenation pattern has been}%
\typeout{** loaded for the language `#1'. Using the pattern for}%
\typeout{** the default language instead.}%
\else
\language=\csname l@#1\endcsname
\fi
#2}}
\providecommand{\BIBdecl}{\relax}
\BIBdecl

\bibitem{Marmanis17}
D.~Marmanis, K.~Schindler, J.~D. Wegner, S.~Galliani, M.~Datcu, and U.~Stilla,
  ``Classification with an edge: Improving semantic image segmentation with
  boundary detection,'' \emph{ISPRS Journal of Photogrammetry and Remote
  Sensing}, vol. 135, no. January, pp. 158--172, 2018.

\bibitem{beyongrgb}
N.~Audebert, B.~L. Saux, and S.~Lef\`evre, ``Beyond {RGB}: Very high resolution
  urban remote sensing with multimodal deep networks,'' \emph{ISPRS Journal of
  Photogrammetry and Remote Sensing}, vol. 140, no. June, pp. 20--32, 2018.

\bibitem{Marcos18}
D.~Marcos, M.~Volpi, B.~Kellenberger, and D.~Tuia, ``Land cover mapping at very
  high resolution with rotation equivariant {CNNs}: Towards small yet accurate
  models,'' \emph{ISPRS Journal of Photogrammetry and Remote Sensing},
  DOI:10.1016/j.isprsjprs.2018.01.021.

\bibitem{mou2018rifcn}
{L. Mou and X. X. Zhu}, ``{RiFCN}: Recurrent network in fully convolutional
  network for semantic segmentation of high resolution remote sensing images,''
  \emph{arXiv:1805.02091}, 2018.

\bibitem{mou2018vehicle}
{L. Mou} and {X. X. Zhu}, ``Vehicle instance segmentation from aerial image and
  video using a multi-task learning residual fully convolutional network,''
  \emph{IEEE Transactions on Geoscience and Remote Sensing}, vol.~56, no.~11,
  pp. 6699--6711, 2018.

\bibitem{7729468}
{L. Mou and X. X. Zhu}, ``Spatiotemporal scene interpretation of space videos
  via deep neural network and tracklet analysis,'' in \emph{IEEE International
  Geoscience and Remote Sensing Symposium (IGARSS)}, 2016.

\bibitem{hsf}
Q.~Li, L.~Mou, Q.~Liu, Y.~Wang, and X.~X. Zhu, ``Hsf-net: Multi-scale deep
  feature embedding for ship detection in optical remote sensing imagery,''
  \emph{IEEE Transactions on Geoscience and Remote Sensing}, 2017.

\bibitem{lucchesi2013applications}
S.~Lucchesi, M.~Giardino, and L.~Perotti, ``Applications of high-resolution
  images and {DTM}s for detailed geomorphological analysis of mountain and
  plain areas of {NW} {I}taly,'' \emph{European Journal of Remote Sensing},
  vol.~46, no.~1, pp. 216--233, 2013.

\bibitem{Mou18}
L.~Mou and X.~X. Zhu, ``{IM2HEIGHT}: Height estimation from single monocular
  imagery via fully residual convolutional-deconvolutional network,''
  \emph{arXiv:1802.10249}, 2018.

\bibitem{weng2018land}
Q.~Weng, Z.~Mao, J.~Lin, and X.~Liao, ``Land-use scene classification based on
  a {CNN} using a constrained extreme learning machine,'' \emph{International
  Journal of Remote Sensing}, vol.~0, no.~0, pp. 1--19, 2018.

\bibitem{cheng2017remote}
G.~Cheng, J.~Han, and X.~Lu, ``Remote sensing image scene classification:
  Benchmark and state of the art,'' \emph{Proceedings of the IEEE}, vol. 105,
  no.~10, pp. 1865--1883, 2017.

\bibitem{zarco2014tree}
P.~Zarco-Tejada, R.~Diaz-Varela, V.~Angileri, and P.~Loudjani, ``Tree height
  quantification using very high resolution imagery acquired from an unmanned
  aerial vehicle ({UAV}) and automatic 3{D} photo-reconstruction methods,''
  \emph{European Journal of Agronomy}, vol.~55, pp. 89--99, 2014.

\bibitem{wen2017semantic}
D.~Wen, X.~Huang, H.~Liu, W.~Liao, and L.~Zhang, ``Semantic classification of
  urban trees using very high resolution satellite imagery,'' \emph{IEEE
  Journal of Selected Topics in Applied Earth Observations and Remote Sensing},
  vol.~10, no.~4, pp. 1413--1424, 2017.

\bibitem{nogueira2017towards}
K.~Nogueira, O.~Penatti, and J.~dos Santos, ``Towards better exploiting
  convolutional neural networks for remote sensing scene classification,''
  \emph{Pattern Recognition}, vol.~61, pp. 539--556, 2017.

\bibitem{zhu2017deep}
X.~X. Zhu, D.~Tuia, L.~Mou, G.~Xia, L.~Zhang, F.~Xu, and F.~Fraundorfer, ``Deep
  learning in remote sensing: A comprehensive review and list of resources,''
  \emph{IEEE Geoscience and Remote Sensing Magazine}, vol.~5, no.~4, pp. 8--36,
  2017.

\bibitem{7358126}
B.~Demir and L.~Bruzzone, ``Histogram-based attribute profiles for
  classification of very high resolution remote sensing images,'' \emph{IEEE
  Transactions on Geoscience and Remote Sensing}, vol.~54, no.~4, pp.
  2096--2107, 2016.

\bibitem{rs71114680}
F.~Hu, G.~Xia, J.~Hu, and L.~Zhang, ``Transferring deep convolutional neural
  networks for the scene classification of high-resolution remote sensing
  imagery,'' \emph{Remote Sensing}, vol.~7, no.~11, pp. 14\,680--14\,707, 2015.

\bibitem{hu2018recent}
F.~Hu, G.~Xia, Y.~W., and Z.~L., ``Recent advances and opportunities in scene
  classification of aerial images with deep models,'' in \emph{IEEE
  International Geoscience and Remote Sensing Symposium (IGARSS)}, 2018.

\bibitem{zhang2015saliency}
F.~Zhang, B.~Du, and L.~Zhang, ``Saliency-guided unsupervised feature learning
  for scene classification,'' \emph{IEEE Transactions on Geoscience and Remote
  Sensing}, vol.~53, no.~4, pp. 2175--2184, 2015.

\bibitem{HUANG2018127}
X.~Huang, H.~Chen, and J.~Gong, ``Angular difference feature extraction for
  urban scene classification using {ZY}-3 multi-angle high-resolution satellite
  imagery,'' \emph{ISPRS Journal of Photogrammetry and Remote Sensing}, vol.
  135, pp. 127 -- 141, 2018.

\bibitem{mou2017multitemporal}
L.~Mou, X.~Zhu, M.~Vakalopoulou, K.~Karantzalos, N.~Paragios, B.~L. Saux,
  G.~Moser, and D.~Tuia, ``Multitemporal very high resolution from space:
  Outcome of the 2016 {IEEE GRSS} data fusion contest,'' \emph{IEEE Journal of
  Selected Topics in Applied Earth Observations and Remote Sensing}, vol.~10,
  no.~8, pp. 3435--3447, 2017.

\bibitem{yang2010bag}
Y.~Yang and S.~Newsam, ``Bag-of-visual-words and spatial extensions for
  land-use classification,'' in \emph{International Conference on Advances in
  Geographic Information Systems (SIGSPATIAL)}, 2010.

\bibitem{xia2017aid}
G.~Xia, J.~Hu, F.~Hu, B.~Shi, X.~Bai, Y.~Zhong, L.~Zhang, and X.~Lu, ``Aid: A
  benchmark data set for performance evaluation of aerial scene
  classification,'' \emph{IEEE Transactions on Geoscience and Remote Sensing},
  2017.

\bibitem{ren2015faster}
S.~Ren, K.~He, R.~Girshick, and J.~Sun, ``Faster {R-CNN}: Towards real-time
  object detection with region proposal networks,'' in \emph{Advances in Neural
  Information Processing Systems}, 2015.

\bibitem{long2015fully}
J.~Long, E.~Shelhamer, and T.~Darrell, ``Fully convolutional networks for
  semantic segmentation,'' in \emph{IEEE Conference on Computer Vision and
  Pattern Recognition (CVPR)}, 2015.

\bibitem{badrinarayanan2015segnet}
V.~Badrinarayanan, A.~Kendall, and R.~Cipolla, ``Segnet: A deep convolutional
  encoder-decoder architecture for image segmentation,''
  \emph{arXiv:1511.00561}, 2015.

\bibitem{viola2001rapid}
P.~Viola and M.~Jones, ``Rapid object detection using a boosted cascade of
  simple features,'' in \emph{IEEE Conference on Computer Vision and Pattern
  Recognition (CVPR)}, 2001.

\bibitem{lin2017feature}
T.~Lin, P.~Doll{\'a}r, R.~Girshick, K.~He, B.~Hariharan, and S.~Belongie,
  ``Feature pyramid networks for object detection,'' in \emph{IEEE conference
  on computer vision and pattern recognition (CVPR)}, 2017.

\bibitem{ren2017object}
S.~Ren, K.~He, R.~Girshick, X.~Zhang, and J.~Sun, ``Object detection networks
  on convolutional feature maps,'' \emph{IEEE transactions on Pattern Analysis
  and Machine Intelligence}, vol.~39, no.~7, pp. 1476--1481, 2017.

\bibitem{karalas2016land}
K.~Karalas, G.~Tsagkatakis, M.~Zervakis, and P.~Tsakalides, ``Land
  classification using remotely sensed data: Going multilabel,'' \emph{IEEE
  Transactions on Geoscience and Remote Sensing}, vol.~54, no.~6, pp.
  3548--3563, 2016.

\bibitem{zeggada2017deep}
A.~Zeggada, F.~Melgani, and Y.~Bazi, ``A deep learning approach to {UAV} image
  multilabeling,'' \emph{IEEE Geoscience and Remote Sensing Letters}, vol.~14,
  no.~5, pp. 694--698, 2017.

\bibitem{koda2018spatial}
S.~Koda, A.~Zeggada, F.~Melgani, and R.~Nishii, ``Spatial and structured {SVM}
  for multilabel image classification,'' \emph{IEEE Transactions on Geoscience
  and Remote Sensing}, pp. 1--13, 2018.

\bibitem{zeggada2018multilabel}
A.~Zeggada, S.~Benbraika, F.~Melgani, and Z.~Mokhtari, ``Multilabel conditional
  random field classification for {UAV} images,'' \emph{IEEE Geoscience and
  Remote Sensing Letters}, vol.~15, no.~3, pp. 399--403, 2018.

\bibitem{hua2019recurrently}
Y.~Hua, L.~Mou, and X.~X. Zhu, ``Recurrently exploring class-wise attention in
  a hybrid convolutional and bidirectional {LSTM} network for multi-label
  aerial image classification,'' \emph{ISPRS Journal of Photogrammetry and
  Remote Sensing}, vol. 149, pp. 188--199, 2019.

\bibitem{shao2013hierarchical}
W.~Shao, W.~Yang, G.~Xia, and G.~Liu, ``A hierarchical scheme of multiple
  feature fusion for high-resolution satellite scene categorization,'' in
  \emph{International Conference on Computer Vision Systems}, 2013.

\bibitem{risojevic2013fusion}
V.~Risojevic and Z.~Babic, ``Fusion of global and local descriptors for remote
  sensing image classification,'' \emph{IEEE Geoscience and Remote Sensing
  Letters}, vol.~10, no.~4, pp. 836--840, 2013.

\bibitem{lowe2004distinctive}
D.~Lowe, ``Distinctive image features from scale-invariant keypoints,''
  \emph{International Journal of Computer Vision}, vol.~60, no.~2, pp. 91--110,
  2004.

\bibitem{7466064}
Q.~Zhu, Y.~Zhong, B.~Zhao, G.~S. Xia, and L.~Zhang, ``Bag-of-visual-words scene
  classifier with local and global features for high spatial resolution remote
  sensing imagery,'' \emph{IEEE Geoscience and Remote Sensing Letters},
  vol.~13, no.~6, pp. 747--751, 2016.

\bibitem{teshome2018novel}
B.~Teshome~Zegeye and B.~Demir, ``A novel active learning technique for
  multi-label remote sensing image scene classification,'' in \emph{Society of
  Photo-Optical Instrumentation Engineers (SPIE) Conference Series}, 2018.

\bibitem{stivaktakis2019deep}
R.~Stivaktakis, G.~Tsagkatakis, and P.~Tsakalides, ``Deep learning for
  multilabel land cover scene categorization using data augmentation,''
  \emph{IEEE Geoscience and Remote Sensing Letters}, 2019.

\bibitem{Demir2019label}
G.~Sumbul and D.~B., ``A {CNN-RNN} framework with a novel patch-based
  multi-attention mechanism for multi-label image classification in remote
  sensing,'' in \emph{IEEE International Geoscience and Remote Sensing
  Symposium (IGARSS)}, 2019.

\bibitem{sumbul2017fine}
G.~Sumbul, R.~Cinbis, and S.~Aksoy, ``Fine-grained object recognition and
  zero-shot learning in remote sensing imagery,'' \emph{IEEE Transactions on
  Geoscience and Remote Sensing}, vol.~56, no.~2, pp. 770--779, 2017.

\bibitem{lee2018multi}
C.~Lee, C.~Yeh, and Y.~F. Wang, ``Multi-label zero-shot learning with
  structured knowledge graphs,'' in \emph{IEEE Conference on Computer Vision
  and Pattern Recognition (CVPR)}, 2018.

\bibitem{santoro2017simple}
A.~Santoro, D.~Raposo, D.~G. Barrett, M.~Malinowski, R.~Pascanu, P.~Battaglia,
  and T.~Lillicrap, ``A simple neural network module for relational
  reasoning,'' in \emph{Advances in Neural Information Processing Systems
  (NIPS)}, 2017.

\bibitem{relationdet}
H.~Hu, J.~Gu, Z.~Zhang, J.~Dai, and Y.~Wei, ``Relation networks for object
  detection,'' in \emph{{IEEE} International Conference on Computer Vision and
  Pattern Recognition (CVPR)}, 2018.

\bibitem{wang2018appearance}
L.~Wang, W.~Li, W.~Li, and L.~Van~Gool, ``Appearance-and-relation networks for
  video classification,'' in \emph{IEEE Conference on Computer Vision and
  Pattern Recognition (CVPR)}, 2018.

\bibitem{zhou2017temporal}
B.~Zhou, A.~Andonian, and A.~Torralba, ``Temporal relational reasoning in
  videos,'' in \emph{European Conference on Computer Vision (ECCV)}, 2018.

\bibitem{mou2019semantic}
L.~Mou, Y.~Hua, and X.~X. Zhu, ``Relation matters: Relational context-aware
  fully convolutional network for semantic segmentation of high resolution
  aerial images,'' \emph{arXiv:1409.1556}, 2019.

\bibitem{jaderberg2015spatial}
M.~Jaderberg, K.~Simonyan, A.~Zisserman, and K.~Kavukcuoglu, ``Spatial
  transformer networks,'' in \emph{Advances in Neural Information Processing
  Systems (NIPS)}, 2015.

\bibitem{chaudhuri2018multilabel}
B.~Chaudhuri, B.~Demir, S.~Chaudhuri, and L.~Bruzzone, ``Multilabel remote
  sensing image retrieval using a semisupervised graph-theoretic method,''
  \emph{IEEE Transactions on Geoscience and Remote Sensing}, vol.~56, no.~2,
  pp. 1144--1158, 2018.

\bibitem{xia2019mining}
F.~Hu, G.~Xia, W.~Yang, and L.~Zhang, ``Mining deep semantic representations
  for scene classification of high-resolution remote sensing imagery,''
  \emph{IEEE Transactions on Big Data}, pp. 1--1, 2019.

\bibitem{zhang2016deep}
L.~Zhang, L.~Zhang, and B.~Du, ``Deep learning for remote sensing data: A
  technical tutorial on the state of the art,'' \emph{IEEE Geoscience and
  Remote Sensing Magazine}, vol.~4, no.~2, pp. 22--40, 2016.

\bibitem{srivastava2019understanding}
S.~Srivastava, J.~E. Vargas-Mu{\~n}oz, and D.~Tuia, ``Understanding urban
  landuse from the above and ground perspectives: A deep learning, multimodal
  solution,'' \emph{Remote Sensing of Environment}, vol. 228, pp. 129--143,
  2019.

\bibitem{wang2019igarss}
W.~Huang, Q.~Wang, and X.~Li, ``Feature sparsity in convolutional neural
  networks for scene classification of remote sensing image,'' in \emph{IEEE
  International Geoscience and Remote Sensing Symposium (IGARSS)}, 2019.

\bibitem{qiu2019local}
{C. Qiu}, {L. Mou}, {M. Schmitt}, and {X. X. Zhu}, ``Local climate zone-based
  urban land cover classification from multi-seasonal {S}entinel-2 images with
  a recurrent residual network,'' \emph{ISPRS Journal of Photogrammetry and
  Remote Sensing}, vol. 154, pp. 151--162, 2019.

\bibitem{qiu2019fusingLetter}
C.~Qiu, L.~Mou, M.~Schmitt, and X.~X. Zhu, ``Fusing multi-seasonal {S}entinel-2
  imagery for urban land cover classification with residual convolutional
  neural networks,'' \emph{IEEE Geoscience and Remote Sensing Letters}, 2019,
  in press.

\bibitem{imagenet_cvpr09}
J.~Deng, W.~Dong, R.~Socher, L.-J. Li, K.~Li, and L.~Fei-Fei, ``Imagenet: A
  large-scale hierarchical image database,'' in \emph{IEEE Conference on
  Computer Vision and Pattern Recognition (CVPR)}, 2009.

\bibitem{nadam2}
T.~Dozat, ``Incorporating {Nesterov} momentum into {Adam},''
  \url{http://cs229.stanford.edu/proj2015/054_report.pdf}, online.

\bibitem{szegedy2015going}
C.~Szegedy, W.~Liu, Y.~Jia, P.~Sermanet, S.~Reed, D.~Anguelov, D.~Erhan,
  V.~Vanhoucke, and A.~Rabinovich, ``Going deeper with convolutions,'' in
  \emph{IEEE Conference on Computer Vision and Pattern Recognition (CVPR)},
  2015.

\bibitem{he2016deep}
K.~He, X.~Zhang, S.~Ren, and J.~Sun, ``Deep residual learning for image
  recognition,'' in \emph{IEEE Conference on Computer Vision and Pattern
  Recognition (CVPR)}, 2016.

\bibitem{wu2016unified}
X.~Wu and Z.~Zhou, ``A unified view of multi-label performance measures,''
  \emph{arXiv:1609.00288}, 2016.

\bibitem{kaggleplanet}
``Planet: Understanding the {A}mazon from space,''
  \url{https://www.kaggle.com/c/planet-understanding-the-amazon-from-space#evaluation},
  online.

\bibitem{tsoumakas2007random}
G.~Tsoumakas and I.~Vlahavas, ``Random k-labelsets: An ensemble method for
  multilabel classification,'' in \emph{European Conference on Machine
  Learning}, 2007.

\bibitem{simonyan2014very}
K.~Simonyan and A.~Zisserman, ``Very deep convolutional networks for
  large-scale image recognition,'' \emph{arXiv:1409.1556}, 2014.

\bibitem{hua2019label}
Y.~Hua, L.~Mou, and X.~X. Zhu, ``Label relation inference for multi-label
  aerial image classification,'' in \emph{IEEE International Geoscience and
  Remote Sensing Symposium (IGARSS)}, 2019.

\end{thebibliography}
\begin{IEEEbiography}[{\includegraphics[width=1in,height=1.25in,clip,keepaspectratio]{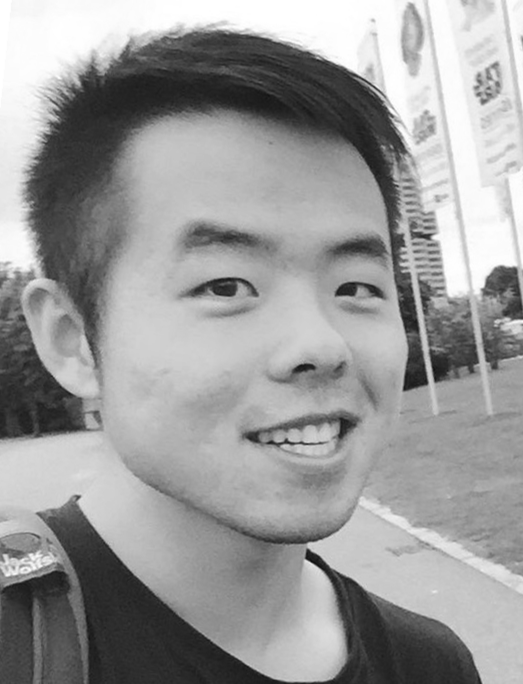}}]{Yuansheng Hua}
(S'18) received the bachelor's degree in remote sensing science and technology from the Wuhan University, Wuhan, China, in 2014, and the master's degree in Earth Oriented Space Science and Technology (ESPACE) from the Technical University of Munich (TUM), Munich, Germany, in 2018. He is currently pursuing the Ph.D. degree with the German Aerospace Center (DLR), Wessling, Germany and the Technical University of Munich (TUM), Munich, Germany. 

In 2019, he was a visiting researcher with the Wageningen University \& Research, Wageningen, Netherlands. His research interests include remote sensing, computer vision, and deep learning, especially their applications in remote sensing.

\end{IEEEbiography}

\begin{IEEEbiography}[{\includegraphics[width=1in,height=1.25in,clip,keepaspectratio]{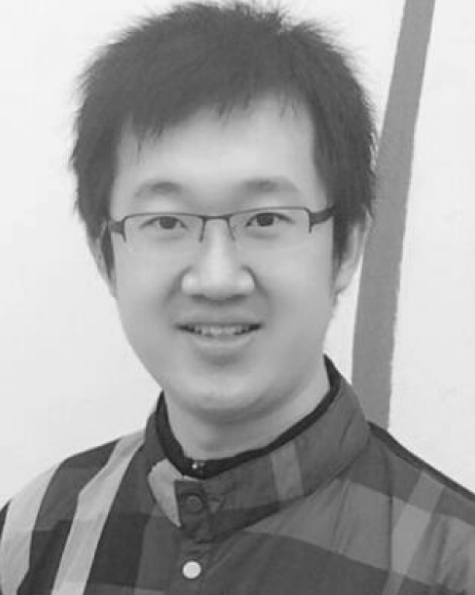}}]{Lichao Mou}
(S'16) received the bachelor's degree in automation from the Xi'an University of Posts and Telecommunications, Xi'an, China, in 2012, and the master's degree in signal and information processing from the University of Chinese Academy of Sciences (UCAS), Beijing, China, in 2015. He is currently pursuing the Ph.D. degree with the German Aerospace Center (DLR), Wessling, Germany, and also with the Technical University of Munich (TUM), Munich, Germany. In 2015, he spent six months at the Computer Vision Group, University of Freiburg, Freiburg im Breisgau, Germany. 

In 2019, he was a Visiting Researcher with the University of Cambridge, Cambridge, U.K. His research interests include remote sensing, computer vision, and machine learning, especially deep networks and their applications in remote sensing.

Mr. Mou was a recipient of the first place in the 2016 IEEE GRSS Data Fusion Contest and finalists for the Best Student Paper Award at the 2017 Joint Urban Remote Sensing Event and the 2019 Joint Urban Remote Sensing Event.
\end{IEEEbiography}


\begin{IEEEbiography}[{\includegraphics[width=1in,height=1.25in,clip,keepaspectratio]{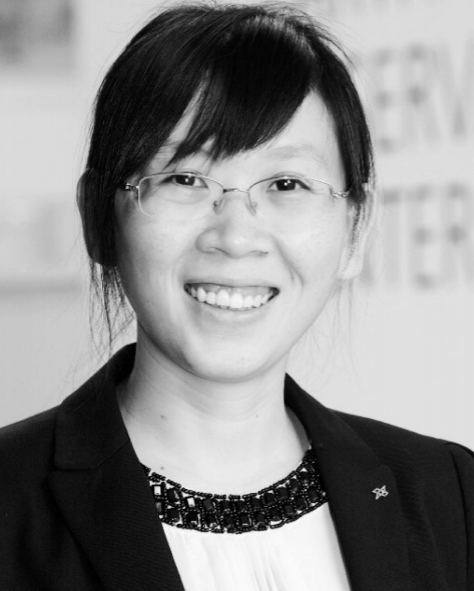}}]{Xiao Xiang Zhu}(S'10--M'12--SM'14) received the Master (M.Sc.) degree, her doctor of engineering (Dr.-Ing.) degree and her “Habilitation” in the field of signal processing from Technical University of Munich (TUM), Munich, Germany, in 2008, 2011 and 2013, respectively.
\par
  She is currently the Professor for Signal Processing in Earth Observation (www.sipeo.bgu.tum.de) at Technical University of Munich (TUM) and German Aerospace Center (DLR); the head of the department ``EO Data Science'' at DLR's Earth Observation Center; and the head of the Helmholtz Young Investigator Group ``SiPEO'' at DLR and TUM. Since 2019, Zhu is co-coordinating the Munich Data Science Research School (www.mu-ds.de). She is also leading the Helmholtz Artificial Intelligence Cooperation Unit (HAICU) -- Research Field ``Aeronautics, Space and Transport". Prof. Zhu was a guest scientist or visiting professor at the Italian National Research Council (CNR-IREA), Naples, Italy, Fudan University, Shanghai, China, the University  of Tokyo, Tokyo, Japan and University of California, Los Angeles, United States in 2009, 2014, 2015 and 2016, respectively. Her main research interests are
  remote sensing and Earth observation, signal processing, machine learning and data science, with a special application focus on global urban mapping.

  Dr. Zhu is a member of young academy (Junge Akademie/Junges Kolleg) at the Berlin-Brandenburg Academy of Sciences and Humanities and the German National  Academy of Sciences Leopoldina and the Bavarian Academy of Sciences and Humanities. She is an associate Editor of IEEE Transactions on Geoscience and Remote Sensing.
\end{IEEEbiography}
\end{document}